\documentclass[twoside,11pt]{article}

\usepackage{hyperref, jair, theapa, rawfonts}
\usepackage{amsmath}
\usepackage{amsthm}
\usepackage{amsfonts}
\usepackage{url}

\usepackage{bm}             
\usepackage{graphics}       
\usepackage{svg}            
\usepackage{comment}        
\usepackage{mathtools}      
\usepackage{mathrsfs}       
\usepackage{listings}       
\usepackage{pgfplots}
\usepackage{multirow}
\usepackage{caption}
\usepackage{subcaption}
\usepackage{booktabs}
\usepackage{algorithm} 
\usepackage{algpseudocode} 
\usepackage{enumitem}   
\usepackage{fixltx2e}   
\usepackage{textcomp}   
\usepackage{cleveref}
\usepackage{wrapfig}

\usepackage{graphicx}        
\usepackage{tikz}            
\usepackage{blindtext}       %
\usepackage[precision=2,unit=mm]{lengthconvert}
\usetikzlibrary{patterns}

\definecolor{color1}{RGB}{141,211,199}
\definecolor{color2}{RGB}{255,255,179}
\definecolor{color3}{RGB}{190,186,218}
\definecolor{color4}{RGB}{251,128,114}
\definecolor{color5}{RGB}{128,177,211}

\newtheorem{innercustomgeneric}{\customgenericname}
\providecommand{\customgenericname}{}
\newcommand{\newcustomtheorem}[2]{%
  \newenvironment{#1}[1]
  {%
   \renewcommand\customgenericname{#2}%
   \renewcommand\theinnercustomgeneric{##1}%
   \innercustomgeneric
  }
  {\endinnercustomgeneric}
}

\newcustomtheorem{customtheorem}{Theorem}
\newcustomtheorem{customlemma}{Lemma}
\newcustomtheorem{customproposition}{Proposition}
\newcustomtheorem{customdefinition}{Definition}
\newcustomtheorem{customexample}{Example}

\definecolor{codegreen}{rgb}{0,0.6,0}
\definecolor{codegray}{rgb}{0.5,0.5,0.5}
\definecolor{codepurple}{rgb}{0.58,0,0.82}
\definecolor{backcolor}{rgb}{0.95,0.95,0.92}
\lstdefinestyle{mystyle}{
    backgroundcolor=\color{backcolor},   
    commentstyle=\color{codegreen},
    numberstyle=\tiny\color{codegray},
    stringstyle=\color{codepurple},
    basicstyle=\ttfamily\footnotesize,
    breakatwhitespace=false,         
    breaklines=true,                 
    captionpos=b,                    
    keepspaces=true,                 
    numbers=left,                    
    numbersep=5pt,                  
    showspaces=false,                
    showstringspaces=false,
    showtabs=false,                  
    tabsize=2
}
\newcommand*{\img}[1]{%
    \raisebox{-.03cm}{%
        \includegraphics[
        height=.32cm,
        width=.32cm,
        keepaspectratio,
        ]{#1}%
    }%
}
\usepackage{pgf} 
\usetikzlibrary[math]
\usepackage{colortbl}
\usepackage{etoolbox}
\definecolor{high}{HTML}{76f013}  
\definecolor{low}{HTML}{3377ff}   

\newcommand{\opacity}{50}         
\newcommand{\minval}{1.0}  
\newcommand{\maxval}{100.0}  
\newcommand{\gradient}[1]{
    \ifdimcomp{#1pt}{>}{\maxval pt}{#1}{
        \ifdimcomp{#1pt}{<}{\minval pt}{#1}{
            \pgfmathparse{int(round(100*(#1/(\maxval-\minval))-(\minval*(100/(\maxval-\minval)))))}
            \xdef\tempa{\pgfmathresult}
            \cellcolor{high!\tempa!low!\opacity} #1
    }}
}

\newcommand{\minvalb}{1}
\newcommand{\maxvalb}{958}
\newcommand{\ctime}[3]{
            \pgfmathparse{100-int(round(((log2((#1*36 + #2*0.6 + #3*0.01)+1)*100 -\minvalb)  / (\maxvalb - \minvalb ))*100))}
            \xdef\tempa{\pgfmathresult}
            \cellcolor{high!\tempa!low!\opacity}
            \ifdimcomp{#1 pt}{>}{0 pt}{#1h:}{}\ifdimcomp{#2 pt}{>}{0 pt}{#2m:}{\ifdimcomp{#1 pt}{>}{0 pt}{0m:}{}}#3s} 

\newcommand{\minvald}{600}
\newcommand{\maxvald}{1250}
\newcommand{\dtime}[4]{
    \pgfmathparse{100-int(round(((log2((#1*864 + #2*36 + #3*0.6 + #4*0.01)+1)*100 -\minvald)  / (\maxvald - \minvald))*100))}
    \xdef\tempa{\pgfmathresult}
    \cellcolor{high!\tempa!low!\opacity}
    \ifdimcomp{#1 pt}{>}{0 pt}{#1d:}{}\ifdimcomp{#2 pt}{>}{0 pt}{#2h:}{}\ifdimcomp{#3 pt}{>}{0 pt}{#3m}{\ifdimcomp{#4 pt}{>}{0 pt}{:#4s}} 
}

\ShortHeadings{Scalable Neural-Probabilistic ASP}{Skryagin, Ochs, Dhami and Kersting}

\firstpageno{1}

\begin{document}

\title{Scalable Neural-Probabilistic Answer Set Programming}

\author{\name Arseny Skryagin$^{1}$ \email arseny.skryagin@cs.tu-darmstadt.de
       \AND
       \name Daniel Ochs$^{1}$ \email daniel.ochs@cs.tu-darmstadt.de
       \AND
       \name Devendra Singh Dhami$^{1,3}$ \email devendra.dhami@cs.tu-darmstadt.de
       \AND
       \name Kristian Kersting$^{1,2,3,4}$ \email kersting@cs.tu-darmstadt.de\\
       \addr $^1$Computer Science Department, TU Darmstadt, Germany\\
        $^2$Centre for Cognitive Science, TU Darmstadt, Germany\\
        $^3$Hessian Center for AI (hessian.AI), Darmstadt, Germany\\
        $^4$German Center for Artificial Intelligence (DFKI), Darmstadt, Germany
}


\maketitle

\begin{abstract}

\noindent The goal of combining the robustness of neural networks and the expressiveness of symbolic methods has rekindled the interest in Neuro-Symbolic AI. 
Deep Probabilistic Programming Languages (DPPLs) have been developed for probabilistic logic programming to be carried out via the probability estimations of deep neural networks.
However, recent SOTA DPPL approaches allow only for limited conditional probabilistic queries and do not offer the power of true joint probability estimation. 
In our work, we propose an easy integration of tractable probabilistic inference within a DPPL.
To this end, we introduce SLASH, a novel DPPL that consists of Neural-Probabilistic Predicates (NPPs) and a logic program, united via answer set programming (ASP). 
NPPs are a novel design principle allowing for combining all deep model types and combinations thereof to be represented as a single probabilistic predicate.
In this context, we introduce a novel $+/-$ notation for answering various types of probabilistic queries by adjusting the atom notations of a predicate.
To scale well, we show how to prune the stochastically insignificant parts of the (ground) program, speeding up reasoning without sacrificing the predictive performance. 
We evaluate SLASH on a variety of different tasks, including the benchmark task of MNIST addition and Visual Question Answering (VQA).
\end{abstract}


\section{Introduction}

\begin{figure}[t]
    \centering
    \includegraphics[scale=0.75]{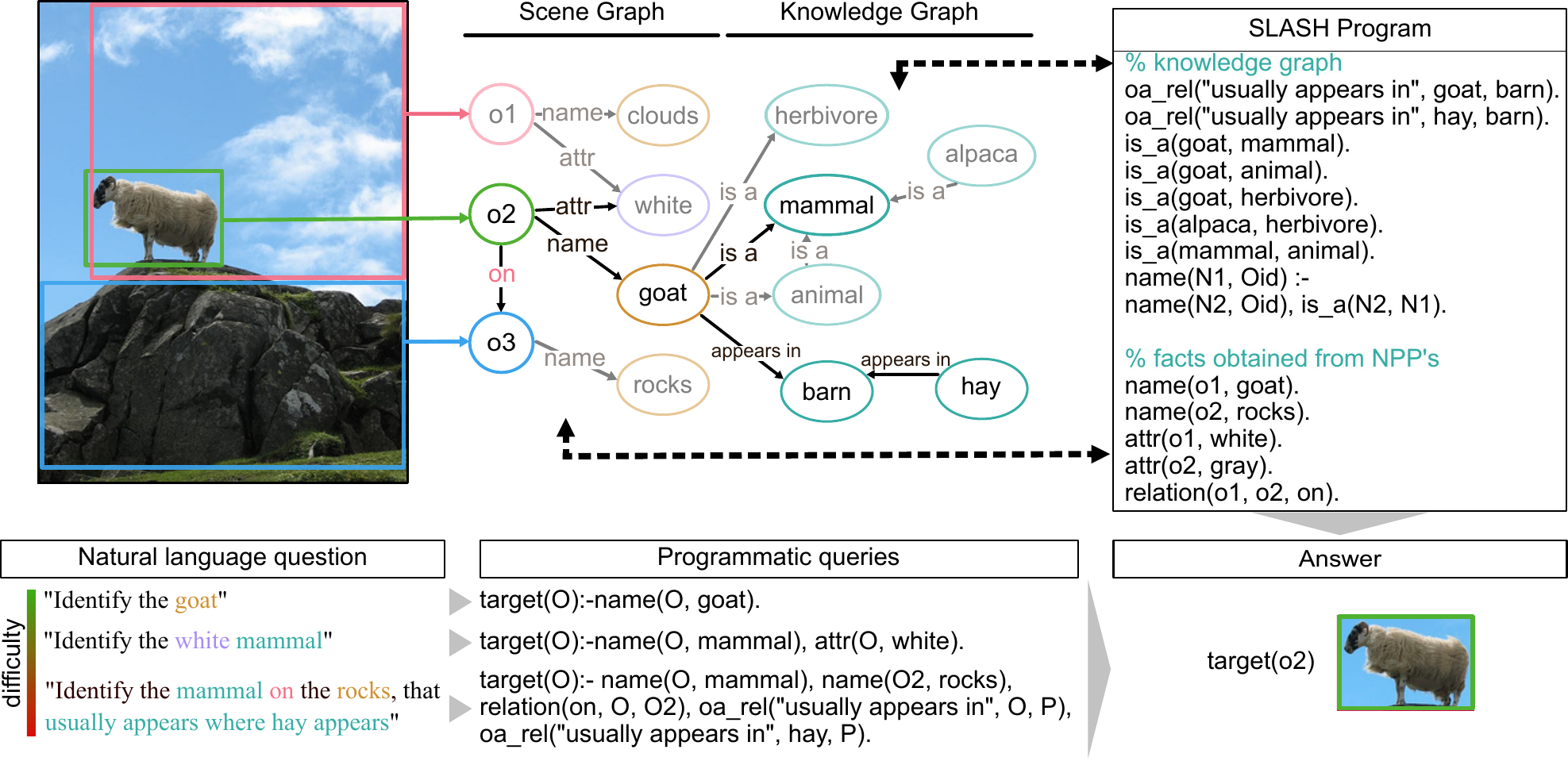}
    \caption{\textbf{The VQA task:} Proposed by \citeA{Scallop},
    given the features and bounding boxes of objects in an image, the goal is to answer a question requiring multi-hop reasoning. 
    A model is learned that predicts a scene graph consisting of names, attributes, and relations.
    Additionally, a fixed knowledge graph is given, extending the scene graph with commonsense knowledge.
    Questions are provided as queries in programmatic form and can vary in complexity, respectively, in the clause length of the query.
    Together, the knowledge- and scene graph are used to infer the correct answer for the query.
    }
    \label{fig:vqa-overview}
\end{figure}
Neuro-symbolic AI approaches to learning \cite{HudsonM19,GarcezGLSST19,JiangA20,GarcezLamb} are on the rise. 
They integrate low-level perception with high-level reasoning by combining data-driven neural modules with logic-based symbolic modules. 
This combination of sub-symbolic and symbolic systems has shown many advantages for various tasks such as VQA and reasoning \cite{yiNSVQA}, concept learning~\cite{MaoNSCL} and improved properties for explainable and revisable models \cite{ciravegna2020human,stammerNeSyXIL}.

Rather than designing specifically tailored neuro-symbolic architectures, where often the neural and symbolic modules are disjoint and trained independently \cite{yiNSVQA,MaoNSCL,stammerNeSyXIL}, deep probabilistic programming languages (DPPLs) provide an exciting alternative \cite{Pyro,Edward,DeepProbLog,NeurASP,Scallop}. 
Specifically, DPPLs integrate neural and symbolic modules via a unifying programming framework with probability estimates acting as the \emph{``glue''} between separate modules, thus allowing for reasoning over noisy, uncertain data and, importantly, joint training of the modules. 
Additionally, prior knowledge and biases in the form of logic rules can easily and explicitly be added to the learning process with DPPLs. 
This stands, in contrast, to specifically tailored, implicit architectural biases of, e.g., purely subsymbolic deep learning approaches. 
Ultimately, DPPLs thereby allow to integrate neural networks easily into downstream logical reasoning tasks.

Recent state-of-the-art DPPLs, such as DeepProbLog \cite{DeepProbLog}, NeurASP \cite{NeurASP} and Scallop \cite{Scallop} allow for conditional class probability estimates as all three works base their probability estimates on neural predicates.
We argue that it is necessary to integrate and process joint probability estimates into DPPLs, to allow for solving a broader range of tasks.
The world is uncertain and it is necessary to reason in settings, e.g., in which variables of an observation might be missing or even manipulated.
\begin{figure}[t]
    \centering
    \includegraphics[width=1\textwidth]{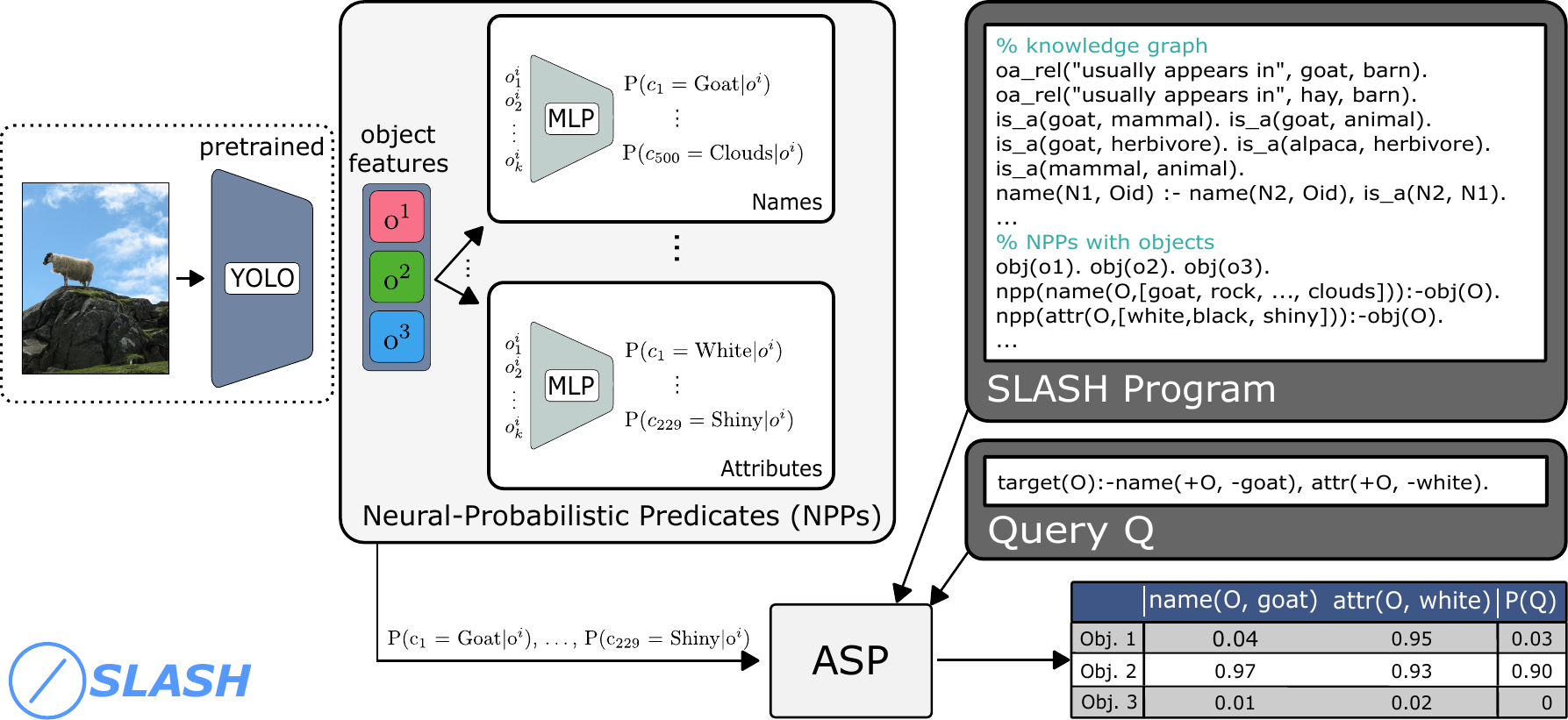}
    \caption{\textbf{VQA task with SLASH:}
    NPPs consist of neural and/or probabilistic circuit modules and can produce task-specific probability estimates. 
    A YOLO Network and MLPs make up the Neural-Probabilistic Predicates for the VQA task.
    In our novel DPPL, SLASH, NPPs are integrated with a logic program via an ASP module to answer logical queries about data samples. 
    Each MLP computes the conditional distribution for classes $c_i$ given the YOLO feature encodings $z_i$ shared across all NPPs, such as names or attributes. 
    The relation's NPP is omitted for simplicity.
    One gets task-related probabilities by sending queries to the NPPs, e.g., conditional probabilities for visual reasoning tasks.
    }
    \label{fig:npp}
    \vspace{-0.2in}
\end{figure}

Hence, we make the following contributions in this work. First, we propose a novel form of predicates for DPPLs, termed Neural Probabilistic Predicates (NPPs, cf. Fig.~\ref{fig:npp}), that allow for task-specific probability queries.
NPPs consist of neural and/or probabilistic circuit (PC) modules and act as a unifying term, encompassing the neural predicates of DeepProbLog, NeurASP and Scallop, as well as purely probabilistic predicates. 
Further, we introduce a much more powerful \emph{``flavor''} of NPPs that consist jointly of neural and PC modules, taking advantage of the power of neural computations together with true density estimation of PCs via tractable probabilistic inference.

Second, having introduced NPPs, we construct SLASH\footnote{Code is available at: \href{https://github.com/ml-research/SLASH}{https://github.com/ml-research/SLASH}}, a novel DPPL, which efficiently combines NPPs with logic programming.
Similar to the punctuation symbol, this can be used to efficiently combine several paradigms into one. 
Specifically, SLASH represents for the first time an efficient programming language that seamlessly integrates probabilistic logic programming with neural representations and tractable probabilistic estimations. 
This allows for the integration of all forms of probability estimations, not just class conditionals, thus extending the important works of \citeA{DeepProbLog}, \citeA{NeurASP} and \citeA{Scallop}.

\begin{wrapfigure}{IR}{0.5\textwidth}
    \centering
    \vspace{-0.2in}
    \includegraphics[scale=0.22]{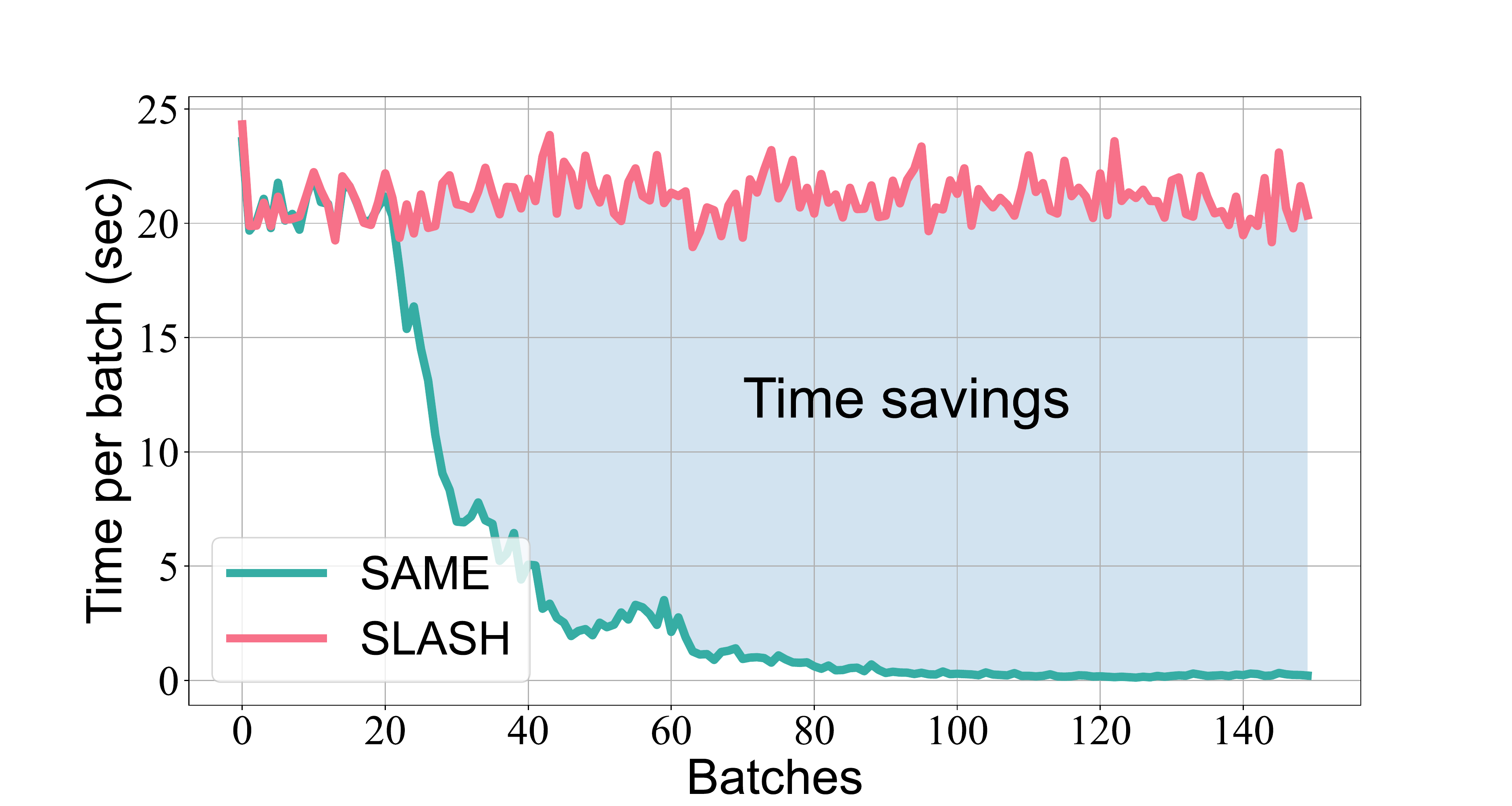}
    \caption{\textbf{SAME helps SLASH to reduce training time:} E.g., on the MNIST T3 task per batch, 
    SAME prunes unlikely outcomes, which reduces the training time of SLASH.}
    \label{fig:same-savings}
\end{wrapfigure}

Third, 
as NPPs become more complex, navigating the solution space becomes more time-consuming. 
To speed up, Scallop \cite{Scallop} used top-k to prune unlikely paths in their proof tree using output probabilities of DNNs.
With PCs (as NPPs) it is, however, difficult to select the correct k.
Instead, we go for top-k\%.
It is based on the observation that for each query there are multiple possible but only one \underline{S}olution th\underline{A}t \underline{M}atch\underline{E}s the data (SAME). 
That is, SAME keeps k\% of SLASH's solutions to compute (probabilistic) answers. 
This greatly speeds up inference, as illustrated in Fig.~\ref{fig:same-savings}.

Moreover, SAME allows SLASH to scale to VQA, as implemented in SLASH in Fig.~\ref{fig:npp}.  
Here, every NPP gets object-detection outputs, in this case from YOLO network~\cite{yolo}, as inputs and produces class conditionals for names, attributes, and relations. 
A user defines a set of statements and rules in the form of a logic program. 
Finally, given the query as in Fig.~\ref{fig:npp}, SLASH gives the expected answer.


The present paper is a significant extension of a previously published conference paper~\cite{SLASH} and presents SAME and how to use it to scale SLASH to VQA.
Further, we extend this previous work with a detailed ablation study:
Empirical results show on the set prediction task carried out on the CLEVR~\cite{CLEVR} dataset as well as on the benchmark task of MNIST-Addition \cite{DeepProbLog}, further presenting the advantages coming with SAME. 


In summary, we make the following contributions:
\begin{itemize}
    \item introduce neural-probabilistic predicates,
    \item efficiently integrate answer set programming (ASP) with probabilistic inference via NPPs within our novel DPPL, SLASH,
    \item introduce SAME to dynamically prune unlikely NPP outcomes, thus allowing a reduction in the complexity of computing potential solutions,
    \item effectively train neural, probabilistic and logic modules within SLASH for complex data structures end-to-end via a simple, single loss term, 
    \item show that the integration of NPPs in SLASH provides various advantages across a variety of tasks and data sets compared to state-of-the-art DPPLs and neural models.
\end{itemize}
These contributions demonstrate the advantage of probabilistic density estimation via NPPs and the benefits of a
``one system -- two approaches'' \cite{bengio2019nips} framework that can successfully be used for performing various tasks and on many data types. 

We proceed as follows. 
First, we introduce NPPs and how they can be queried via the $+/-$ notation. 
Next, SLASH programs are presented with the corresponding semantics and parameter learning. 
Afterwards, we discover SAME using top-k\%.
Before concluding, we support our findings with experimental evaluation.

\begin{figure}[t!]
    \centering
        \centering
        \includegraphics[width=0.98\textwidth]{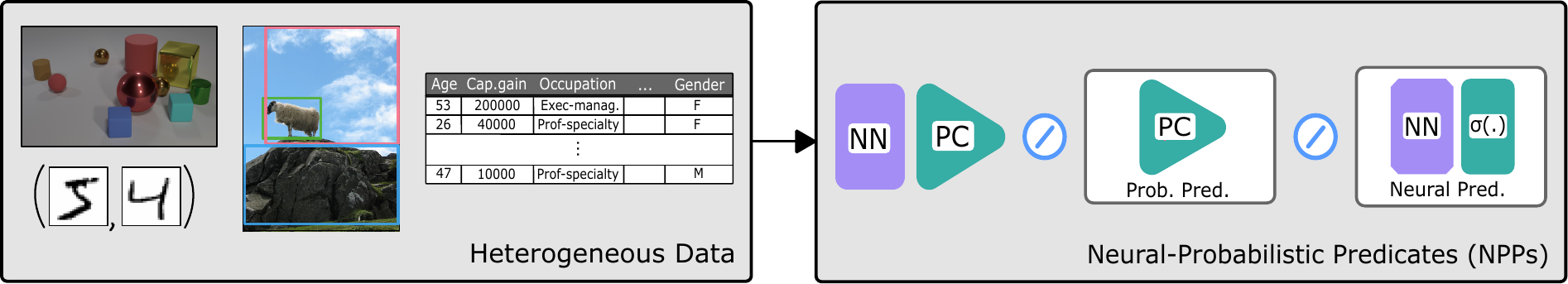}
    \caption{\textbf{NPPs come in various flavors:}
    Depending on the data set and underlying task, SLASH requires a suitable Neural-Probabilistic Predicate (NPP) to compute query-dependent probability estimates. 
    NPPs can be composed of neural and probabilistic modules, or (depicted via the slash symbol) only one of these two.}
    \label{fig:npp-building-blocks}
\end{figure}

\section{SLASH through NPPs and vice versa}  
We begin this section by first introducing the novel neural probabilistic predicates (NPPs) framework. 
After this, we introduce our DPPL, SLASH, which easily integrates NPPs via ASP with logic programming and end this section with the learning procedure in SLASH, allowing us to train all modules via a joint loss term. 

\subsection{Neural-Probabilistic Predicates and Rules}
Previous DPPLs, DeepProbLog \cite{DeepProbLog} and NeurASP \cite{NeurASP}, introduced the \textit{Neural Predicate} as an annotated-disjunction or as a propositional atom, respectively, to acquire conditional class probabilities, $P(C|X)$, via the softmax function at the output of an arbitrary DNN.
As mentioned in the introduction, this approach has certain limitations concerning inference capabilities. To resolve this issue, we introduce \textit{Neural-Probabilisitic Predicates} (NPPs).

Formally, we denote with
\begin{equation}\label{eq:def_NPP}
    npp\left(h(x), \left[v_1,\ldots,v_n\right]\right)
\end{equation}
a Neural-Probabilistic Predicate $h$.
Thereby, (i) \textit{npp} is a reserved word to label a NPP, (ii) \textit{h} a symbolic name of either a PC, NN or a joint of a PC and NN (\textit{cf.} Fig.~\ref{fig:npp-building-blocks})(right), e.g., \textit{name} is the name of a NPP of Fig.~\ref{fig:npp} (right, `SLASH Program'-block). 
Additionally, (iii) \textit{$x$} denotes a ``term'' and (iv) $v_1,\ldots,v_n$ are the $n$ possible outcomes of $h$. 
For example, the placeholders for \textit{name} are the names of an object (\textit{goat, rocks, \ldots, clouds}).

A NPP abbreviates a rule of the form $c=v$ with $c\in \{ h(x)\}$ and $v\in\{v_1,\ldots,v_n\}$. 
Furthermore, we denote with $\Pi{\mbox{\it npp}}$ a set of NPPs of the form stated in (Eq.~\eqref{eq:def_NPP}) and $r^{\mbox{\it npp}}$ the set of all rules $c=v$ of one NPP, which denotes the possible outcomes, obtained from a NPP in $\Pi{\mbox{\it npp}}$, e.g., $r^{name} = \{c=\mbox{\it Goat}, c=Rocks,\ldots, c=Clouds\}$ for the example depicted in Fig.~\ref{fig:npp}.
Rules in the following form   
\begin{equation}\label{eq:npp_rule}
    npp\left(h(x),\left[v_1,\ldots,v_n\right]\right) \leftarrow \textit{Body}
\end{equation}
are used as an abbreviation for application to multiple entities, e.g., multiple object features plus bounding boxes for the VQA task (\textit{cf.} Fig.~\ref{fig:npp}).
Hereby, \textit{Body} of the rule is identified by $\top$ (tautology, true) or $\bot$ (contradiction, false) during grounding. 
Rules of the form \textit{Head $\leftarrow$ Body} with $r^{\mbox{\it npp}}$ appearing in \textit{Head} are prohibited for $\Pi{\mbox{\it npp}}$. 

In this work, we use NPPs that contain probabilistic circuits, which allow for tractable density estimation and modelling of joint probabilities.
The term probabilistic circuit (PC) \cite{ProbCirc20} represents a unifying framework encompassing all computational graphs that encode probability distributions and guarantee tractable probabilistic modelling.
These include Sum-Product Networks (SPNs) \cite{SPNs}, which are deep mixture models represented via a rooted directed acyclic graph with a recursively defined structure.
In this way, with PCs it is possible to answer a much richer set of probabilistic queries, i.e. $P(X,C)$, $P(X|C)$, $P(C|X)$ and $P(C)$. 

In addition to NPPs based purely on PCs, we introduce the arguably more interesting type of NPP that combines a neural module with a PC. 
Hereby, the neural module learns to map the raw input data into an optimal latent representation. 
The PC, in turn, learns to model the joint distribution of these latent variables and produces the final probability estimates. 
This type of NPP nicely combines the representational power of neural networks with the advantages of PCs in probability estimation and query flexibility. 
These combined NPPs can be partially pretrained or trained end-to-end. 
In the VQA example, we utilize a pretrained YOLO network with an MLP predicting class conditional probabilities. In object-centric learning, we train a slot-attention module and PCs over the latent representations end-to-end (see Sec.~\ref{sec:object-centric-learning}).

To make the different probabilistic queries distinguishable in a SLASH program, we follow mode declarations used in inductive logic programming (ILP), and denote the input variable with $+$ and the output variable with $-$.
E.g., within the example of VQA (\textit{cf.} Fig.~\ref{fig:npp}, `Query Q' (right)), with the query $name(+X, -C)$ one is asking for $P(C|X)$ with C being the class and X the object features. 
If we chose a PC as the underlying network (c.f. Sec.~\ref{sec:object-centric-learning} and \ref{mnist-exp}) we can model the joint distribution $P(X, C)$. 
Similarly, with $name(-X,+C)$ one is asking for $P(X|C)$ and, finally, with $name(+X,+C)$ for $P(X,C)$. 
In the case when no data is available, i.e, $name(-X,-C)$, we are querying for the prior $P(C)$.

To summarize, a NPP can consist of neural and/or probabilistic modules and produces query-dependent probability estimates. 
Due to the flexibility of its definition, the term NPP contains the predicates of previous works \cite{DeepProbLog,NeurASP}, but also more interesting predicates discussed above. 
The specific ``flavor'' of a NPP should be chosen depending on what type of probability estimation is required (\textit{cf.} Fig~\ref{fig:npp-building-blocks}).


\subsection{SLASH: a novel DPPL for integrating NPPs}
Now we have everything together to introduce SLASH, a novel DPPL that efficiently integrates NPPs with logic programming.

\subsubsection{SLASH Language and Semantics}
We continue in the pipeline, Fig.~\ref{fig:npp}, on how to use the probability estimates of NPPs for answering logical queries, and begin by formally defining a \textbf{SLASH program}.
A SLASH program $\Pi$ is the union of $\Pi^{\mbox{\it asp}}$, $\Pi^{\mbox{\it npp}}$. 
Therewith, $\Pi^{\mbox{\it asp}}$ is the set of propositional rules (standard rules from ASP-Core-2 \cite{ASP-Core-2}), and $\Pi^{\mbox{\it npp}}$ is a set of Neural-Probabilistic Predicates of the form stated in Eq.~\eqref{eq:def_NPP}.

Similar to NeurASP, SLASH requires ASP and, as such, adopts its \textbf{syntax} for the most part, which includes neural probabilistic rules as defined in Eq.~\eqref{eq:npp_rule}.
Compared to ProbLog, ASP rarely goes into an infinite loop (cf. Chapter 2.9 of~\citeA{lifschitz2019answer} for a simple code example leading to infinite loops with ProbLog) and is therefore preferable as a backbone. 
To illustrate, let us revisit the example of VQA as in Fig.~\ref{fig:vqa-overview}. 
A YOLO network detected three objects $\mbox{\it o1}$, $\mbox{\it o2}$ and $\mbox{\it o3}$ in the image. 
The task is to $\mbox{\it name}$ each of the objects as either $\mbox{\it goat}$, $\mbox{\it rock}$, $\ldots$  or $\mbox{\it clouds}$.
The overall $\mbox{\it target}$ here is to find an object $\mbox{\it goat}$: 
\begin{align*}
    & \mbox{\it obj}(\mbox{\it o1}). \quad \mbox{\it obj}(\mbox{\it o2}). \quad \mbox{\it obj}(\mbox{\it o3}).\\
    & \mbox{\it npp}(\mbox{\it name}(O,[\mbox{\it goat}, \mbox{\it rock}, \ldots, \mbox{\it clouds}])) \leftarrow \mbox{\it obj}(O). \\
    & \mbox{\it target}(O) \leftarrow \mbox{\it name}(+O, \mbox{\it -goat}).
\end{align*}
Fig.~\ref{fig:vqa-overview} presents one further SLASH program for the task of VQA, exemplifying a set of propositional rules and neural predicates. 

Now, let us define the \textbf{semantics} of SLASH.
To this end, we show how to integrate NPPs into an ASP-compatible form to obtain the \textbf{success probability for a query} given all potential solutions, i.e., stable models. 
A \textbf{query} is an ASP constraint of the form $\leftarrow\textit{Body}$, i.e., it is a headless rule. 
To translate the program $\Pi$, the rules (Eq.~\eqref{eq:npp_rule}) will be rewritten as follows:
\begin{equation}\label{eq:rewrite_NPP}
    1\{h(x)=v_1;\ldots;h(x)=v_n\}1  
\end{equation}
The ASP-solver should understand this as ``Pick exactly one rule from the set''. 
After the translation is done, we can ask an ASP-solver for the solutions for $\Pi$.

Next, let us assume that we have a query $Q$ at hand for which we want to compute the probability; keep in mind that NPPs introduce random choices.
Since all the potential solutions $I\models Q$ for the query $Q$ are mutually exclusive, there are possible worlds, the probability $ P_{\Pi}(Q)$ of $Q$ is the sum of probabilities $P_{\Pi}(I)$ of each single solution, i.e., stable model of $Q$:
 \begin{equation}\label{eq:def_query_prob}
        P_{\Pi}(Q) := \sum_{I\models Q} P_{\Pi}(I)\;\text{.} 
 \end{equation}
So, we are left with computing the probability $P_{\Pi}(I)$ of a single solution $I$. Here, only the NPPs are contributing to the probability; all other atoms are simply true and have the probability $1$. The (ground) NPPs, however, are also independent of each other. Consequently, for each object $c$ and random choice $v$, we can multiply together the probabilities of $c=v$ and normalize by the number of objects $c$:
\begin{equation}\label{eq:def_pot_sol_prob}
    P_{\Pi}(I) =
    \begin{cases}
        \frac{\prod_{c=v\in I|_{r^{\it npp}}}P_{\Pi}(c=v)}{|I|_{r^{\it npp}}|}, & \text{if } I \text{ is pot. sol. of } \Pi\text{,} \\
        0, & \text{otherwise.}
    \end{cases}
\end{equation}
where $I|_{r^{\mbox{\it npp}}}$ is the subset of ground NPP, $r^{\mbox{\it npp}}$ in the solution $I$, $r^{\mbox{\it npp}}\subseteq I$.

With the success probability $P_{\Pi}(Q)$ of a single query at hand, the success probability of a set of queries $\mathbf{Q}$ can naturally be written as 
\begin{equation}\label{eq:def_query_set_prob}
        P_{\Pi}\left(\mathbf{Q}\right) := \prod_{i=1}^{l} P_{\Pi}(Q_i) = \prod_{i=1}^{l}\sum_{I\models Q_i} P_{\Pi}(I)\text{}
\end{equation}
since they are independent of each other. 
With the semantics at hand, we are ready to learn the parameters of SLASH programs. 

\subsubsection{Parameter Learning in SLASH}
To estimate the parameters  $\bm{\theta}$ of a SLASH program $\Pi(\bm{\theta})$, we are following the {\it learning from entailment} setting, as also used for DeepProblog~\cite{DeepProbLog}. 
That is, we estimate $\bm{\theta}$ from a set $\mathbf{Q}$ of positive examples only, i.e., 
each training example is a logical query that is known to be true in the SLASH program $\Pi(\bm{\theta})$. 
Thereby, $\Pi(\bm{\theta}) = \Pi^{\mbox{\it asp}}(\bm{\theta}) \cup \Pi^{\mbox{\it npp}}(\bm{\theta})$ holds. 
Since $\Pi^{\mbox{\it asp}}(\bm{\theta})$ has no weighted rules, i.e., $P_{\Pi^{\mbox{\it asp}}(\bm{\theta})}=1$, we want to find optimal parameters $\bm{\theta}$ for $r^{\mbox{\it npp}}$, i.e., the optimal NPP parameters.

To achieve parameter learning in SLASH, we employ the loss function, which is additive.
The first part is the {\it entailment} loss, i.e., the NPPs are fixed, and we maximize the success probability of the query set $\bm{Q}$.
The second part concerns the NPPs (neural networks/ probabilistic circuits) only. 
So, we want to maximize the probability given the data while the ``logical'' part is fixed.
Thus, the loss function takes the following form
\begin{equation}\label{eq:loss_total}
    L_{SLASH} = L_{ENT} + L_{NPP}\;,
\end{equation}
and we seek to minimize the loss, e.g., by running coordinate descent. Let us begin with the NPP loss.



\textbf{NPP loss} -- The aim of this loss function is to maximize the joint probability of $P^{(X_{\mathbf{Q}}, C)}_{\mathbf{\theta}}(x_{\mathbf{Q}})$.
To omit possibly vanishing values, we apply $\log(\cdot)$ instead and define
\begin{align}\label{eq:loss_NPP}
    L_{NPP} & := \log\left(P^{(X_{\mathbf{Q}}, C)}_{\mathbf{\theta}}(x_{\mathbf{Q}})\right) = \sum_{i=1}^{l}\log(P^{(X_{\mathbf{Q}}, C)}_{\mathbf{\theta}}(x_{Q_i}))\;,
\end{align}
whereby
\begin{itemize}
    \item $X_{\mathbf{Q}}$ are the random variables modeling the training set $X$ associated with the set of the queries $\mathbf{Q}$,
    \item $x_{\mathbf{Q}}$ are realizations of $X_{\mathbf{Q}}$ associated with $\mathbf{Q}$ as well,
    \item $P^{(X_{\mathbf{Q}},C)}(x_{\mathbf{Q}})$ is the probability of the realizations $x_{\mathbf{Q}}$ estimated by the NPP modelling the joint over the set $X_{\mathbf{Q}}$ and $C$ -- the set of classes (the domain of the NPP cf. Eq.~\eqref{eq:def_NPP}),
    \item and $\mathbf{\theta}$ is the parameter set associated with the NPP.
\end{itemize}
Additionally, we derive the derivative of the NPP loss function, which will be called upon during the training with coordinate descent.
Formally, we write
\begin{equation}\label{eq:grad_loss_NPP}
    \frac{\partial}{\partial\mathbf{\theta}} L_{NPP} = \sum_{i=1}^{l}  \dfrac{1}{P_{\mathbf{\theta}}^{(X_{\mathbf{Q}},C)}(x_{Q_i})} \cdot \frac{\partial}{\partial \mathbf{\theta}}\left(P_{\mathbf{\theta}}^{(X_{\mathbf{Q}},C)}(x_{Q_i})\right)\;.
\end{equation}
Now, we fix the NPPs and proceed with the entailment loss.

\textbf{Entailment loss -- } We begin with $L_{ENT}$ and more concrete with Eq.~\eqref{eq:def_query_set_prob}.
Dealing with probabilities, we might end up with vanishing small values due to the product. 
To resolve it, we apply $\log(\cdot)$ on both sides of the equation and obtain
\begin{equation}\label{eq:def_query_set_prob_log}
    \log(P_{\Pi(\bm{\theta})}(\mathbf{Q}) = \sum_{i=1}^{l}\log\left(\sum_{I\models Q_i}P_{\Pi(\bm{\theta})}(I)\right).
\end{equation}
Since our goal is to give the ``feedback'' of the success probability Eq.~\eqref{eq:def_query_set_prob_log} to NPPs, we multiply it with the log-probabilities of NPPs, so that the result lands in the same space  
\begin{equation}\label{eq:feedback}
    \log(P_{\Pi(\theta)}(\mathbf{Q}))\cdot\log\left(P^{(X_{\mathbf{Q}}, C)}(x_{\mathbf{Q}})\right)\text{.}
\end{equation}
More precisely, we want Eq.~\eqref{eq:feedback} to resonate with every class encoded as a possible outcome $v_j$ as defined in Eq.~\eqref{eq:def_NPP} and with every query $Q_i$ from $\mathbf{Q}$
\begin{align}\label{eq:loss_entailment}
    & \sum_{i=1}^{l}\sum_{j=1}^{n}\log(P_{\Pi(\theta)}(Q_{i}))\cdot\log\left(P^{(\mathbf{Q}, C_j)}(x_{Q_{i}})\right) = \nonumber \\
    & \log LH\left(\log(P_{\Pi(\mathbf{\theta)}}(\mathbf{Q})), P^{(X_{\mathbf{Q}}, C)}(x_{\mathbf{Q}}) \right) =: L_{ENT}\text{.}            
\end{align}
In the above, we used the definition of the log-likelihood loss to compound every single query $i$ and outcome $j$ to the single term of the {\it entailment loss}.
We remark that the defined loss function is true regardless of the NPP's form (NN with Softmax, PC or PC jointly with NN).
The only difference will be the second term, e.g., $P^{(C|X_{\mathbf{Q}})}(x_{\mathbf{Q}})$ or $P^{(X_{\mathbf{Q}}|C)}(x_{\mathbf{Q}}))$ depending on the NPP and task. 
This loss function aims at maximizing the estimated success probability for a set of Queries.
However, for NPPs to notice the ``feedback'' Eq.~\eqref{eq:feedback} we must make Eq.~\eqref{eq:def_query_set_prob_log} relatable towards the log-probabilities of NPPs.

\textbf{Gradients of the entailment loss} -- Particularly, we denote the vector $\log\left(P^{(X_{\mathbf{Q}}, C)}(x_{\mathbf{Q}})\right)$ as $\mathbf{p}$ and denote with $\tfrac{\partial\log\left( P_{\Pi(\bm{\theta})}(\mathbf{Q})\right)}{\partial \mathbf{p}}$ the communication bridge between $\log(P_{\Pi(\bm{\theta})}(\mathbf{Q}))$ and $\mathbf{p}$.
So, we write
\begin{align}\label{eq:grad_via_backprop}
    \sum_{i=1}^{l} \frac{\partial\log\left( P_{\Pi(\bm{\theta})}(Q_i)\right)}{\partial \mathbf{p}} \times \frac{\partial \mathbf{p}}{\partial \bm{\theta}} =
    \sum_{i=1}^{l} \frac{\partial\log\left( P_{\Pi(\bm{\theta})}(Q_i)\right)}{\partial \bm{\theta}} \text{,}
\end{align}
reminding ourselves that $\tfrac{\partial \mathbf{p}}{\partial\bm{\theta}}$ can be computed as usual via backward propagation through the NPPs.
If within the SLASH program, $\Pi(\bm{\theta})$, the NPP forwards the data tensor through a NN first, i.e., the NPP models a joint over the NN's output variables by a PC, then we rewrite Eq.~\eqref{eq:grad_via_backprop} to 
\begin{equation}\label{eq:stacked_NPP_grad}
    \sum_{i=1}^{l} \frac{\partial\log\left( P_{\Pi(\bm{\theta})}(Q_i)\right)}{\partial \mathbf{p}} \times \frac{\partial \mathbf{p}}{\partial \bm{\theta}} \times \frac{\partial\bm{\theta}}{\partial\bm{\gamma}} = \sum_{i=1}^{l} \frac{\partial\log\left( P_{\Pi(\bm{\theta})}(Q_i)\right)}{\partial \bm{\theta}}\text{.}
\end{equation}
Thereby, $\bm{\gamma}$ is the set of the NN's parameters and again, we compute $\tfrac{\partial\bm{\theta}}{\partial\bm{\gamma}}$ via backward propagation.

Now, $\tfrac{\partial\log\left( P_{\Pi(\bm{\theta})}(Q)\right)}{\partial\mathbf{p}}$ is left to be determined.
Thus, following the definition from NeurASP \cite{NeurASP}, we write
\begin{equation}\label{eq:NeurASP:grad}
    \dfrac{\partial\log\left( P_{\Pi(\bm{\theta})}(Q)\right)}{\partial\mathbf{p}} := 
    \left(\sum\limits_{\substack{I: I\models Q, \\ I\models c=v }} \dfrac{P_{\Pi(\bm{\theta})}(I)}{P_{\Pi(\bm{\theta})}(c=v)}
    - \sum\limits_{\substack{I,v': I\models Q, \\ I\models c=v', v\ne v' }} \dfrac{P_{\Pi(\bm{\theta})}(I)}{P_{\Pi(\bm{\theta})}(c=v')}\right)
    \cdot
    \dfrac{1}{P_{\Pi}(Q)}\text{.} 
\end{equation} 

\begin{algorithm}[t]
	\caption{Gradient computation}\label{gradient-pseudo} 
	\begin{algorithmic}[1]
        \Require $P_{\Pi}(c=v_j),j=1,\ldots,n$, set of $I\models Q$ 
            \State $P_{\Pi}(I)\gets \text{compute\_pot\_sol\_prob}(I)$ \textcolor{codegreen}{~\# cf. Eq. \eqref{eq:def_pot_sol_prob} }
            \State $P_{\Pi}(Q)\gets \text{compute\_query\_prob}(P_{\Pi}(I))$ \textcolor{codegreen}{~\# $(:= \gamma)$ Normalization, cf. Eq~\eqref{eq:def_query_prob}}
            \State $\mbox{grads} \gets \emptyset $
                \For {every $c=v_j$}\textcolor{codegreen}{~\# cf. Eq.~\eqref{eq:rewrite_NPP}}
                \State $\mbox{grad} \gets 0$
                    \For {\text{every pot. sol.} $I$}
                        \If{$I\models c=v$} 
                           \State $\mbox{grad} \gets  \mbox{grad} + (P_{\Pi}(I)/P_{\Pi}(c=v))$ \textcolor{codegreen}{~\# $(:= \alpha)$ Reward}
                        \Else
                            \State $\mbox{grad} \gets \mbox{grad} - (P_{\Pi}(I)/P_{\Pi}(c=v'))$\textcolor{codegreen}{~\# $(:= \beta)$ Penalty} 
                            \EndIf
                    \EndFor
                    \State $\mbox{grads} \gets \text{append}(\mbox{grads}, \mbox{grad} / P_{\Pi}(Q))$
			\EndFor\\
        \Return $\mbox{grads}$
	\end{algorithmic} 
\end{algorithm}

\noindent Reading the right-hand side of this definition we recognize the three terms: inside the parentheses, from the reward ($=:\alpha$) is the penalty ($=:\beta$) subtracted, and the result is normalized with the probability of the query ($:=\gamma$), cf. Eq.~\eqref{eq:def_query_prob}.
As the definition stipulates running inference is enough and so having defined the gradients in Eq.~\eqref{eq:NeurASP:grad}, it is of prime interest to have insights into their values. 
The following theorem shows the limit of the gradient vector.

\begin{customtheorem}{1 (Gradients' Limit)}\label{thm:grad-conv}
    Let $\Pi(\bm{\theta})$ be a fixed program with a given query $Q$. Further, $m$ denotes a training iteration, then the following holds for $\tfrac{\partial\log\left( P_{\Pi(\bm{\theta})}(Q)\right)}{\partial\mathbf{p}}$ as defined in \eqref{eq:NeurASP:grad}: 
    \[\left( \dfrac{\partial\log\left( P_{\Pi(\bm{\theta})}(Q)\right)}{\partial\mathbf{p}}\right)_{m} \xrightarrow{m\rightarrow\infty} \left(-1,\ldots,-1,\underbrace{1}_{j},-1,\ldots,-1\right).\]
    Thereby, the index $j$ corresponds to $c=v$ and any other to $c=v', v\ne v'$.
\end{customtheorem}
\begin{proof}
    W.l.o.g, we assume the program $\Pi(\theta)$ to entail a single NPP, and it can be called upon more than once in a single rule $r_{npp}$.
    Besides, a NPP can converge ``perfectly'', i.e., $P_{\Pi(\bm{\theta})}(c=v)=1$ and $P_{\Pi(\bm{\theta})}(c=v')=0$ for $v\ne v'$.
    To answer the question of how such limit values are possible in the first place, we make the observation on the right-hand side of \eqref{eq:NeurASP:grad}, 
    \begin{align*}
        &\sum\limits_{\substack{I: I\models Q \\ I\models c=v }} \tfrac{P_{\Pi(\bm{\theta})}(I)}{P_{\Pi(\bm{\theta})}(c=v)}\cdot P_{\Pi(\bm{\theta})}(c=v)
        + \sum\limits_{\substack{I,v': I\models Q \\ I\models c=v', v\ne v' }} \tfrac{P_{\Pi(\bm{\theta})}(I)}{P_{\Pi(\bm{\theta})}(c=v')}\cdot P_{\Pi(\bm{\theta})}(c=v') = \\
        & \sum\limits_{\substack{I: I\models Q \\ I\models c=v }} P_{\Pi(\bm{\theta})}(I) + \sum\limits_{\substack{I,v': I\models Q \\ I\models c=v', v\ne v' }}P_{\Pi(\bm{\theta})}(I) =
        \sum\limits_{I: I\models Q}P_{\Pi(\bm{\theta})}(I) = P_{\Pi}(Q).
    \end{align*}
    As reward, penalty, and normalization constant are defined before the theorem:
    \[
        \alpha:=\sum\limits_{\substack{I: I\models Q \\ I\models c=v }} \tfrac{P_{\Pi(\bm{\theta})}(I)}{P_{\Pi(\bm{\theta})}(c=v)},\quad
        \beta:= \sum\limits_{\substack{I,v': I\models Q \\ I\models c=v', v\ne v' }} \tfrac{P_{\Pi(\bm{\theta})}(I)}{P_{\Pi(\bm{\theta})}(c=v')},\quad
        \gamma:= \sum\limits_{I: I\models Q}P_{\Pi(\bm{\theta})}(I),
    \]
    we conclude that
    \[\label{eq:grad_ineq}
        \gamma \ge \alpha - \beta \qquad\text{and}\qquad \alpha + \beta \ge \gamma. \tag{*}
    \]
    Now, we consider the following case discrimination based on the training iteration k:
    \begin{enumerate}[label=(\roman*)]
        \item \textit{For $m=0$:}
        At the start of the training, the probabilities of $n$ outcomes are either uniformly distributed (the probability of each outcome $P_{\Pi(\bm{\theta})}(c=v_{j}), j\in{1,\ldots,n}$ is the same) or there are small numerical differences.
        Here, we consider the first possibility and the latter is identical to (ii). 
        Since the probability for each outcome is the same value, we conclude due to \eqref{eq:grad_ineq} that $\alpha = \beta$ and $\tfrac{\alpha - \beta}{\gamma} = 0$ for the index $j$.
        In case that the same NPP being called upon multiple times, an ASP solver will derive potential solutions without consideration of symmetries.
        Consequentially, we have to swap the numerical values obtained in the previous case for $\alpha$ and $\beta$. 
        Nonetheless, we obtain the same gradient value for such a case, i.e., $0$.
        For the rest of the indices, $\alpha=0$ and all values being pulled to $\beta$. Hence, we obtain $-\tfrac{2\beta}{\gamma}$. I.e., for the rest of the indices, the gradient value is negative.
        \item \textit{For any $1\le m<\infty$:}
        At the index $j$ - since $0<\gamma\le 1$ holds, and due to \eqref{eq:grad_ineq} we get 
        \[\label{eq:grad_upper_bound}
            \alpha - \beta \le \gamma\; |: \gamma \quad \Longleftrightarrow \quad \frac{\alpha - \beta}{\gamma} \le 1. \tag{**}
        \]
        In case that the same NPP being called upon multiple times, we multiply \eqref{eq:grad_upper_bound} with $-1$ and obtain
        \[
            -1 \le \frac{\beta - \alpha}{\gamma}.
        \]
        For all other indices, we have $-\tfrac{\beta}{\gamma}$. 
        If $|\beta|>\gamma$, then $-\tfrac{\beta}{\gamma} < -1$ occurs as well.
        \item \textit{For $m=\infty$}, if the NPP fully converged, then we have two cases to distinguish: the index $j$ and all other entries of the gradient's vector.
        Particularly, we know from Eq.~\eqref{eq:def_query_prob} and ~\eqref{eq:def_pot_sol_prob} that $\gamma$ is equal to $1$. 
        Thus, we can focus entirely on $\alpha - \beta$. 
        I.e., we conclude the entries of gradient's vector to be calculated as $1-0=1$ for the index $j$ and $0-1=-1$ otherwise.
    \end{enumerate} 
\end{proof}

Following the theorem, the training is done by the principle ``winner takes all'' if there are more than two NPP's outcomes, and ``zero-sum game'' otherwise. 
Hence, we are left with the sign function of the gradient vector, and the convergence in itself, can be thought of as a gradient clipping.
The results presented by \citeA{NeurASP} show that this works on some problems with little or no loss of accuracy, cf.~\cite{Seide}.
Extrapolating from the gradient's vector limit, we see only one outcome to be rewarded, and so only one of the set of all potential solutions matching the data per NPP's call.
This observation is the heart of the next section and will be discussed in detail.

Now, it is of great interest to derive the gradients of the {\it entailment loss} $L_{ENT}$ \eqref{eq:loss_entailment} so that the expression $\tfrac{\partial\log\left( P_{\Pi(\bm{\theta})}(Q_i)\right)}{\partial \mathbf{p}} \times \tfrac{\partial \mathbf{p}}{\partial \bm{\theta}}$ from the left-hand side of Eq.~\eqref{eq:grad_via_backprop} becomes attainable for back-propagation.
For the very sole purpose, we formulate the 
\begin{customtheorem}{2 (Gradient with respect to entailment loss)}\label{thm:entailment_loss_estimation}
    The average derivative of the logical entailment loss function $L_{ENT}$ defined in Eq.~\eqref{eq:loss_entailment} can be estimated as follows
    \begin{equation*}
      \frac{1}{l}\frac{\partial}{\partial \mathbf{p}}L_{ENT} \ge \frac{1}{l}\sum_{i=1}^{l} \frac{\partial \log(P_{\Pi(\theta)}(Q_i))}{\partial \mathbf{p}}\cdot \log(P^{(X_{\mathbf{Q}}, C)}(x_{Q_i})).
    \end{equation*}
\end{customtheorem}
\begin{proof}
    We begin with the definition of the \textbf{cross-entropy} for two vectors $y_i$ and $\hat{y}_i$:
    \begin{align*}
        -H(y_i,\hat{y}_i) := -\sum_{j=1}^{n} y_{ij}\cdot\log\left(\frac{1}{\hat{y}_{ij}}\right) 
         = -\sum_{j=1}^{n} \left(y_{ij}\cdot\underbrace{\log(1)}_{=0} - y_{ij}\cdot\log(\hat{y}_{ij})\right) 
         = \sum_{j=1}^{n} y_{ij}\cdot\log(\hat{y}_{ij})\text{.}
    \end{align*}
    Hereafter, we substitute
    \begin{equation*}
        y_i = \log(P_{\Pi(\theta)}(Q_i)) \qquad\text{ and }\qquad \hat{y}_i = P^{(X_{\mathbf{Q}}, C)}(x_{Q_i}),
    \end{equation*}
    and thus obtain $-H(y_i,\hat{y}_i) = $
    \begin{equation}\label{loss_derivation:1}
        -H\left(\log(P_{\Pi(\theta)}(Q_i)), P^{(X_{\mathbf{Q}}, C)}(x_{Q_i})\right)
        = \sum_{j=1}^{n}\log(P_{\Pi(\theta)}(Q_{i}))\cdot\log\left(P^{(X_{\mathbf{Q}}, C_j)}(x_{Q_{i}})\right)\text{.}
    \end{equation}
    We remark that $n$ represent the number of classes defined in the domain of an NPP. 
    Now, we differentiate the equation (\ref{loss_derivation:1}) with the respect to $\mathbf{p}$ depicted as in Eq.~\eqref{eq:NeurASP:grad} to be the label of the probability of an atom $c = v_j$ in $r^{\mbox{\it npp}}$, denoting $P_{\Pi(\bm\theta)}(c = v_j)$.
    Since differentiation is a linear operation, the product rule is applicable directly: 
    \begin{align}
        -\dfrac{\partial}{\partial\mathbf{p}} H\left(y_i, \hat{y}_i\right) = 
        \sum_{j=1}^{n}\Bigg[ & \frac{\partial \log(P_{\Pi(\theta)}(Q_{i}))}{\partial\mathbf{p}}\cdot \log\left(P^{(X_{\mathbf{Q}}, C_j)}(x_{Q_{i}})\right) + \nonumber \\
        & \log(P_{\Pi(\theta)}(Q_{i}))\cdot\frac{\partial \log\left(P^{(X_{\mathbf{Q}}, C_j)}(x_{Q_{i}})\right)}{\partial\mathbf{p}}\Bigg]\text{.}
    \end{align}
    We want to avoid considering the latter term of $\log(P_{\Pi(\theta)}(Q_i))\cdot\tfrac{\partial \log\left(P^{(X_{\mathbf{Q}}, C_j)}(x_{Q_i})\right)}{\partial\mathbf{p}}$ because it represents the rescaling and to keep the first since SLASH procure $\tfrac{\partial \log(P_{\Pi(\theta)}(Q_i))}{\partial\mathbf{p}}$ following  Eq.~\eqref{eq:NeurASP:grad}.
    To achieve this, we estimate equation from above downwards as 
    \begin{align}
        -\frac{\partial}{\partial\mathbf{p}}  H\left(y_i, \hat{y}_i\right) \ge
        \sum_{j=1}^{n}\frac{\partial \log(P_{\Pi(\theta)}(Q_{i}))}{\partial\mathbf{p}}\cdot \log\left(P^{(X_{\mathbf{Q}}, C_j)}(x_{Q_{i}})\right)\text{.} \label{loss_derivation:3}
    \end{align}
    Furthermore, under i.i.d assumption we obtain from the definition of likelihood
    \begin{equation*}
        LH(y,\hat{y}) = \prod_{i=1}^{l}LH(y_i,\hat{y}_i)\text{,}
    \end{equation*}
    and following the negative likelihood coupled with the knowledge that the log-likelihood of $y_i$ is the log of a particular entry of $\hat{y}_i$
    \begin{align*}
        L_{ENT} & = \log LH(y,\hat{y}) = \sum_{i=1}^{l}\log LH(y_i,\hat{y}_i) = \sum_{i=1}^{l}\sum_{j=1}^{n}y_{ij}\cdot\log(\hat{y}_{ij})\\ 
        &  = \sum_{i=1}^{l}\left[\sum_{j=1}^{n}y_{ij}\cdot\log(\hat{y}_{ij})\right] = -\sum_{i=1}^{l}H(y_i,\hat{y}_i) \text{.}
    \end{align*}
    Finally, we obtain the following estimate applying inequality \eqref{loss_derivation:3}
    \begin{align*}
        \frac{1}{l}\frac{\partial}{\partial\mathbf{p}}L_{ENT}  = -\frac{1}{l} \sum_{i=1}^{l}\frac{\partial}{\partial\mathbf{p}}H(y_i,\hat{y}_i) \ge \frac{1}{l} \sum_{i=1}^{l}\frac{\partial \log(P_{\Pi(\theta)}(Q_i))}{\partial\mathbf{p}}\cdot \log\left(P^{(X_{\mathbf{Q}}, C)}(x_{Q_i})\right)\text{.}
    \end{align*}
    Also, we note that the mathematical transformations listed above hold for any type of NPP and the task dependent queries (NN with Softmax, PC or PC jointly with NN).
    The only difference will be the second term, i.e., $\log(P^{(C|X_{\mathbf{Q}})}(x_{Q_{i}}))$ or $\log(P^{(X_{\mathbf{Q}}|C)}(x_{Q_{i}}))$ depending on the NPP and task.
    The NPP in a form of a single PC modeling the joint over $X_{\mathbf{Q}}$ and $C$ was depicted to be the example.

\end{proof}
In summary, we have covered the parameter learning within SLASH since the gradients for both $L_{NPP}$ and $L_{ENT}$ are derived, and thus, know the gradients of $L_{SLASH}$. 
Importantly, with the learning schema described above, rather than requiring a novel loss function for each individual task and data set, with SLASH, it is now possible to simply incorporate specific task and data requirements into the logic program. 
The training loss, however, remains the same. 

\section{Scaling SLASH with SAME}
\label{sec:same}
In the following, we focus on the potential solutions $I\models Q$.
Namely, according to Eq.~\eqref{eq:def_query_prob} we know that the probability of a query is the sum of the probabilities of all potential solutions. 
However, the question remains, how many of them match the data $x$ belonging to the query $Q$?
Discussing Thm.~\ref{thm:grad-conv}, we saw that gradients converge to reward only one outcome $v_j$.

We examine this observation on the digit addition task as it was originally proposed by \citeA{DeepProbLog}.
The goal is to train an NPP to recognize digits given the sum of the two. 
For example,  consider the query $\mbox{\it sum2}$(\img{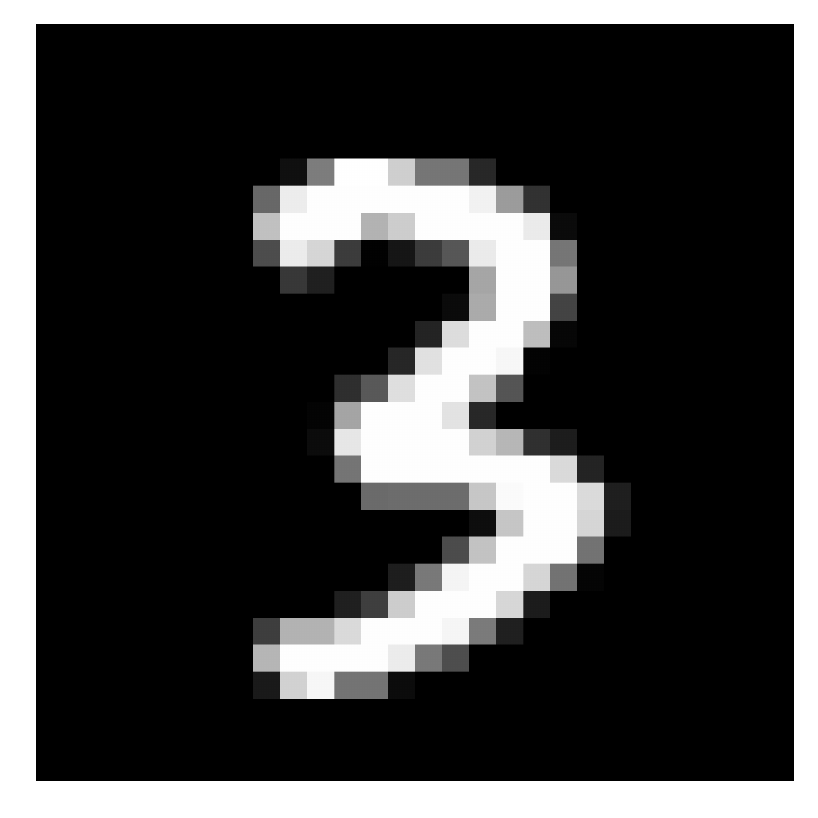},\img{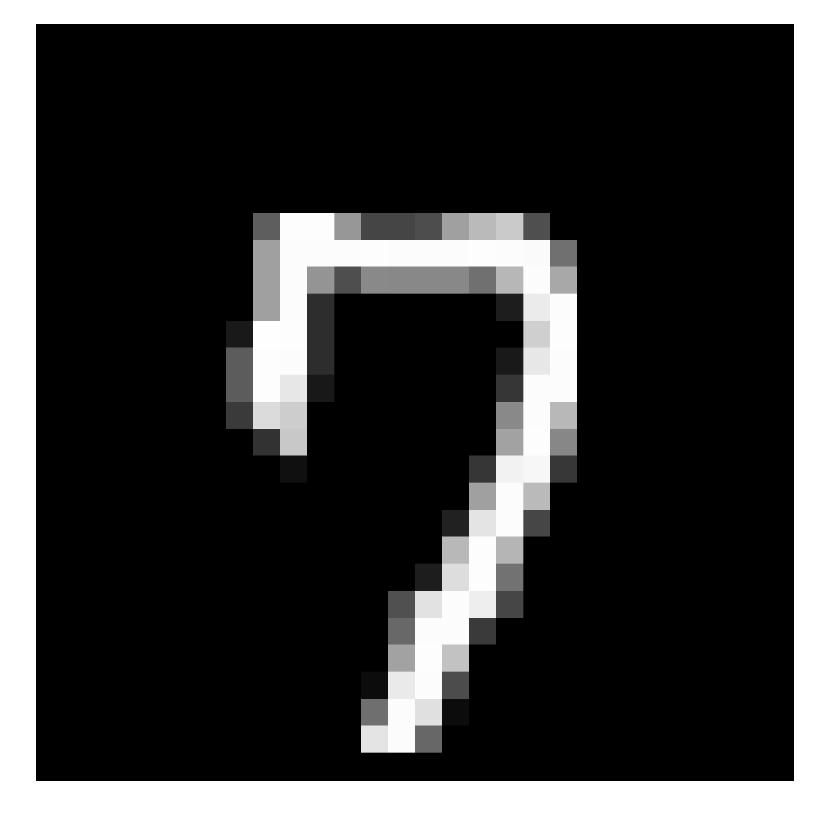},10).
The potential solutions for this query are: 
\begin{align*}
    &\mbox{\it sum2}(1,9,10), \mbox{\it sum2}(2,8,10), \mbox{\it sum2}(3,7,10), \mbox{\it sum2}(4,6,10), \mbox{\it sum2}(5,5,10),\\
    &\mbox{\it sum2}(6,4,10), \mbox{\it sum2}(7,3,10), \mbox{\it sum2}(8,2,10), \mbox{\it sum2}(9,1,10).
\end{align*}
From the above, only $\mbox{\it sum2}$(3,7,10) corresponds to the given data. 
This means we always generate all potential solutions for the given query, although only one corresponds to the data assigned to the query.
In the following, we formulate SAME (Solution thAt MatchEs the data), a technique to focus only on such potential solutions over time and dynamically reduces the computation time spent deriving all potential solutions. 
In the following, we abbreviate with SAME the usage of SAME within SLASH. 

For every query $Q$ SLASH answers, it produces a set of all potential solutions $\mathcal{I}$.
With the growing size of NPP's domain $n$, grows $\mathcal{I}$ substantially.
Having multiple NPPs with considerable domain size, we might end-up with a computationally infeasible set $\mathcal{I}$ to obtain.

During the training, we observe that the probability distribution $P_{\Pi}(c=v_i)$
as defined below Eq.~\eqref{eq:def_NPP} becomes skewed independent of the chosen inference type through $+/-$ notation for every data entry $x$ assigned to the query $Q$.
I.e., with the progressing training's iteration, fewer and fewer NPP's outcomes $v_i$ concentrate the vast majority of the critical mass, or more formally
\begin{equation}\label{eq:def_SAME_threshold}
    \sum_{j}P_{\Pi}(c=v_j) \le t.
\end{equation}
Thereby, $t$ represents some preset threshold of, e.g., 99\%. 
Furthermore, we know that at the beginning of the training $P_{\Pi}(c=v_{j})=\tfrac{1}{n}$ applies for all $v_{i}$, $1\le j \le n$.
Thus, the disjunction in Eq.~\eqref{eq:rewrite_NPP} consists of $n$ elements and $\mathcal{I}_{0} = \mathcal{I}$.
Repeatedly applying Eq.~\eqref{eq:def_SAME_threshold}, we expect the aforementioned disjunction to entail fewer elements with every further training iteration. 
I.e., there exists an order such that
\begin{equation}\label{eq:skewed_prob}
    \sum_{e}P_{\Pi}(c=v_e) < \sum_{j}P_{\Pi}(c=v_j)\quad\text{ with }\quad j\neq e, 1\leq j,e\leq n  
\end{equation}
\begin{algorithm}[t]
	\caption{Potential Solutions with SAME}
    \label{same-pseudo} 
	\begin{algorithmic}[1]
        \Require $P_{\Pi}(c=v_j),j=1,\ldots,n$, $t$, $\Pi^{\mbox{\it asp}}$
            \State $\Pi^{\mbox{\it npp}} \gets \emptyset $ \textcolor{codegreen}{~ initialize the set of NPP, cf. Eq.~\eqref{eq:rewrite_NPP}}
			\For {every $c=v_j$}
                \State $\mbox{prob}_{\mbox{sort}} \gets$  sort$\left(P_{\Pi}(c=v_j)\right)$
                \State \textcolor{codegreen}{~\# add indices of outcomes $v_{j}$ until  $\sum_{j}P_{\Pi}(c=v_j) \le t$ with SAME}
                \State $\mbox{idx} \gets \text{get\_idx}(\mbox{prob}_{\mbox{sort}},t)$
                \State \textcolor{codegreen}{~\# Then truncate disjunction $1\{h(x)=v_1;\ldots;h(x)=v_n\}1$ accordingly}
                \State  $\Pi^{\mbox{\it npp}} \gets \text{extend}(\Pi^{\mbox{\it npp}},\text{get\_disj}(\mbox{idx}))$    
			\EndFor
        \State $\Pi =  \Pi^{\mbox{\it npp}} \cup \Pi^{\mbox{\it asp}}$
        \State $I \gets \text{asp\_solver}(\Pi)$\\
        \Return $I$
	\end{algorithmic} 
\end{algorithm}
\noindent We refer to the Algorithm~\ref{same-pseudo} of SAME in pseudocode form as a summary of the considerations made. 
It depicts how SAME is used when computing all potential solutions.
Consequentially,
\begin{equation}\label{eq:descendent_subset_seq}
    \mathcal{I}_0 \supseteq \mathcal{I}_1 \supseteq \mathcal{I}_2 \supseteq \ldots \supseteq \mathcal{I}_m
\end{equation}
is a formal description of our expectations, and $|\mathcal{I}_m| = 1$ for $m\rightarrow\infty$.
I.e., among all potential solutions, there exists only one potential solution aligning the data with the query. 
Together with Eq.~\eqref{eq:loss_total} and \eqref{eq:def_SAME_threshold} we formulate the following theorem. 
\begin{customtheorem}{3 (Convergence of SAME)}
    Eq.~\eqref{eq:descendent_subset_seq} holds. 
\end{customtheorem}
\begin{proof}
    We follow the principal of contraposition. 
    W.l.o.g., there exists $m\in\mathbb{N}$ such that $\mathcal{I}_{m} \subseteq \mathcal{I}_{m+1}$ holds and not in contrary $\mathcal{I}_{m} \supseteq \mathcal{I}_{m+1}$.
    I.e., the set of the potential solutions in an $m+1$ iteration entails more elements than the set in the previous iteration, or more formally $|\mathcal{I}_{m+1}|\ge |\mathcal{I}_{m}|$.
    Furthermore, if this tendency remains to be true for every subsequent iteration, we obtain
    \begin{equation*}
        \mathcal{I}_{m} \subseteq \mathcal{I}_{m+1} \subseteq \mathcal{I}_{m+2} \subseteq \ldots \subseteq \mathcal{I}_{m+s}\quad\text{ with }\quad s \in\mathbb{N}.
    \end{equation*}
    Since any $\mathcal{I}_{m+s}$ cannot entail more entries than the set of all potential solutions, we conclude
    \begin{equation}\label{eq:limit_set_sequence}
        \mathcal{I}_{m+s} \rightarrow \mathcal{I}=\mathcal{I}_{0}\quad\text{ for }\quad s\rightarrow\infty.
    \end{equation}
    We have shown that SAME would add more and more potentials solutions until it reaches the upper bound of all potential solutions which coincide with the query $Q$.
    All of the above is true for any arbitrary $m\in\mathbb{N}$, thereby completing the proof.
\end{proof}
In the next section, we will provide empirical evidence for the advantages SAME's utilization brings.

\section{Experimental Evaluations}
\label{sec:exp_eval}
Previously, we showed that the main advantage of SLASH lies in the efficient integration of any combination of neural, probabilistic and symbolic computations. 
This work extends these findings with new experimental evaluations for SLASH with SAME. 
In particular, we show how SAME is essential for using SLASH for VQA.
Afterward, we conduct an ablation study to evaluate the advantages coming from this combination. 
For this, we revisit the MNIST addition as conducted by \citeA{Scallop} and the set prediction task as proposed by \citeA{SlotAttention}.
For all experiments, we use top-99\% to cover most critical mass of NPPs.

In the ablation study experiments, we present the average over five runs with different random seeds for parameter initialization.
For VQA experiments, we used the same single seed to initialize the NPP's parameters following the setting of \citeA{Scallop}.
We refer to App.~\ref{app:slash-programs} for each experiment's SLASH program, including queries, and App.~\ref{app:experimental-details} for a detailed description of hyperparameters and further experimental details. 
\subsection{Visual Question Answering}
\label{vqa-exp}

\begin{figure}[t]
    \centering
    \begin{subfigure}{.5\textwidth}
      \centering
        \includegraphics[scale=0.6]{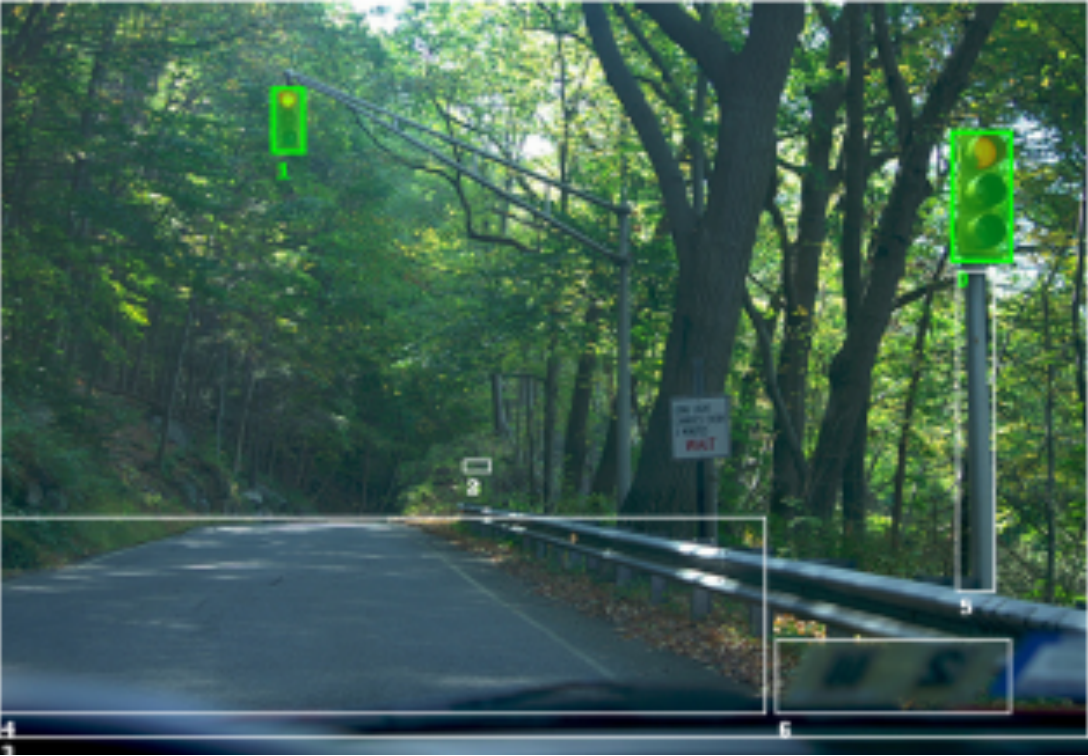}
        \mbox{\it target}(\mbox{\it O0}) :- \mbox{\it name}(\mbox{\it +O0}, \mbox{\it -N0}), \mbox{\it oa\_rel}(\mbox{\it is\_used\_for}, \mbox{\it N0}, \mbox{\it controlling\_flows\_of\_traffic}).
        \caption{Traffic lights example from VQAR C\textsubscript{2}}    
        \label{fig:vqar-traffic}
      \label{fig:sub1}
    \end{subfigure}%
    \begin{subfigure}{.5\textwidth}
        \centering
        \includegraphics[scale=0.6]{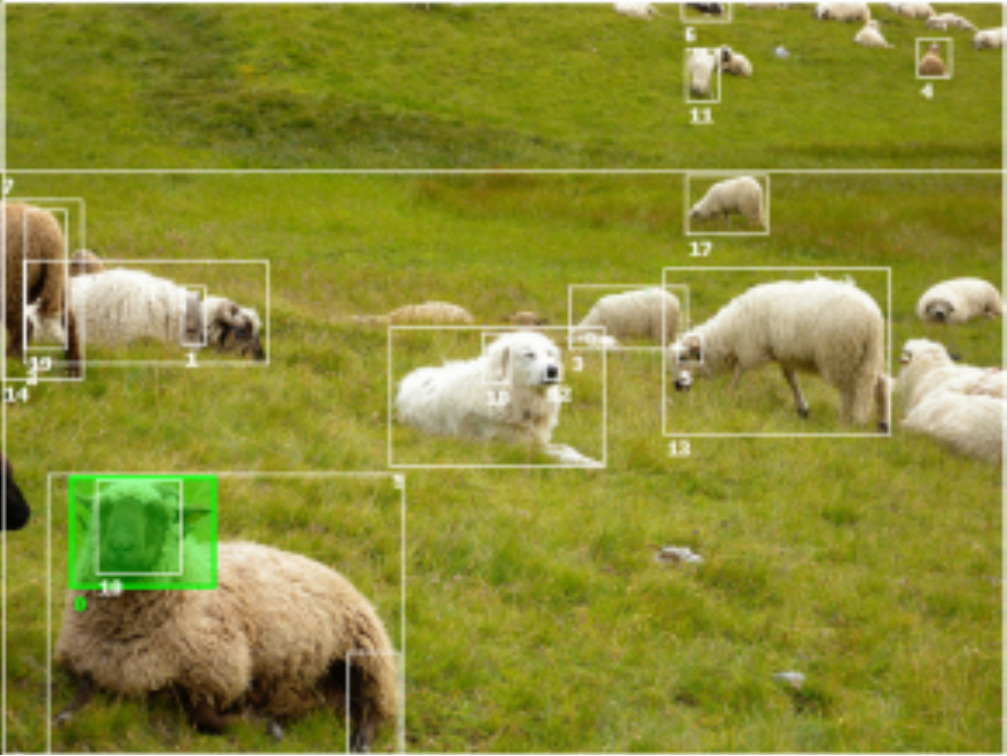}
        \mbox{\it target}(\mbox{\it O2}) :- \mbox{\it relation}(\mbox{\it +O2}, \mbox{\it +O1},  \mbox{\it -of}), \mbox{\it name}(\mbox{\it +O1}, \mbox{\it -animal}), \mbox{\it name}(\mbox{\it O1}, \mbox{\it object}), \mbox{\it relation}(\mbox{\it +O0}, \mbox{\it +O1},  \mbox{\it -of}).    
        \caption{Flock of sheep example from VQAR C\textsubscript{5}}
        \label{fig:vqar-sheep}
      
    \end{subfigure}    
    \caption{\textbf{VQAR example images and programmatic queries:} Bounding boxes are produced by a YOLO network
    and answer objects are marked with green. 
    On the right, the $\mbox{\it name}(\mbox{\it O1}, \mbox{\it object})$ predicate is not annotated with the $+/-$ notation and has to be derived via the knowledge graph.}
    \label{fig:vqar-example}
\end{figure}
In VQA, a model should produce answers to questions about visual scenes. 
These questions require a range of capabilities to infer the correct answer. 
For example, to answer the question ``How many red objects are in the scene?'' a model has to be able to detect and count red objects.
In this experiment, we show how SLASH can be applied to VQA to answer questions that require reasoning. 

As of now, few works approach VQA using logic-based DPPLs \cite{vienne-nesy,Scallop}.
Both of these works open up the question of how ASP can be used in an end-to-end trainable setting; for example, questions about scenes from real-world images, such as in the VQAR dataset proposed by \citeA{Scallop}.
We will now investigate how to apply SLASH to the VQAR dataset.

\textbf{Task Description -- } The VQAR dataset consists of 80.178 real-world images.
Fig.~\ref{fig:vqa-overview} gives an overview of the task. 
Each image was fed through a pretrained YOLO Network to obtain bounding boxes and feature maps for recognized objects.
Each image has a scene graph (SG), which can have 500 object names, 609 attributes and 229 object relations among the objects.
All images share a knowledge graph (KG) encoding 3.387 entries as tuples and triplets, and six rules to traverse. 
Both graphs are represented in the form of a logic program. 
There are 4M programmatic queries and answer pairs encoding object identification questions. 
The queries' difficulty varies, ranging from two to six occurring clauses (C\textsubscript{2} to C\textsubscript{6}), and for each image, ten query answer pairs exist for each clause length. 
Fig.~\ref{fig:vqar-example} depicts two examples of VQAR from C\textsubscript{2} and C\textsubscript{5}. 
In Fig.~\ref{fig:vqa-overview}, next to the programmatic queries are their corresponding natural language questions to be found.
Similarly, as \citeA{Scallop}, we argue that this work focuses on enabling reasoning for VQA, and as such, we use the programmatic form as input. 
Some works, such as \cite{yiNSVQA}, translate from natural language to programmatic queries.
We leave this for future work.

\textbf{Approach by SAME -- } The task is formulated as a multi-label classification task. 
The feature maps, bounding boxes, the entire knowledge graph and the programmatic query serve as input to predict the objects that answer the programmatic query. 
Fig.~\ref{fig:npp} shows the SLASH pipeline for VQA.
In our setup, three MLP classifiers are used as NPPs to predict names, attributes, and relations and are trained end-to-end.
All three are of the same architecture (cf. App.~\ref{app:model-details}) as defined by \citeA{Scallop}.
The NPPs outcomes form the scene graph and build the SLASH program with the KG and the query.
The VQA task, in itself, exposes the limits of DPPLs without approximate reasoning.
The complexity of the real world is so high that the complete enumeration of all proofs/models is beyond reach.
We use a combination of SAME, CLINGO's show statements and iterative solving to deal with the complexity of the task.
We refer the interested reader to App.~\ref{app:experimental-details-vqa}, where we look in-depth into our program encoding.
In the following, we compare SLASH using SAME with Scallop.

\begin{figure}[t]
\centering
    \begin{subfigure}[b]{0.43\textwidth}
        \centering
        \includegraphics[scale=0.21]{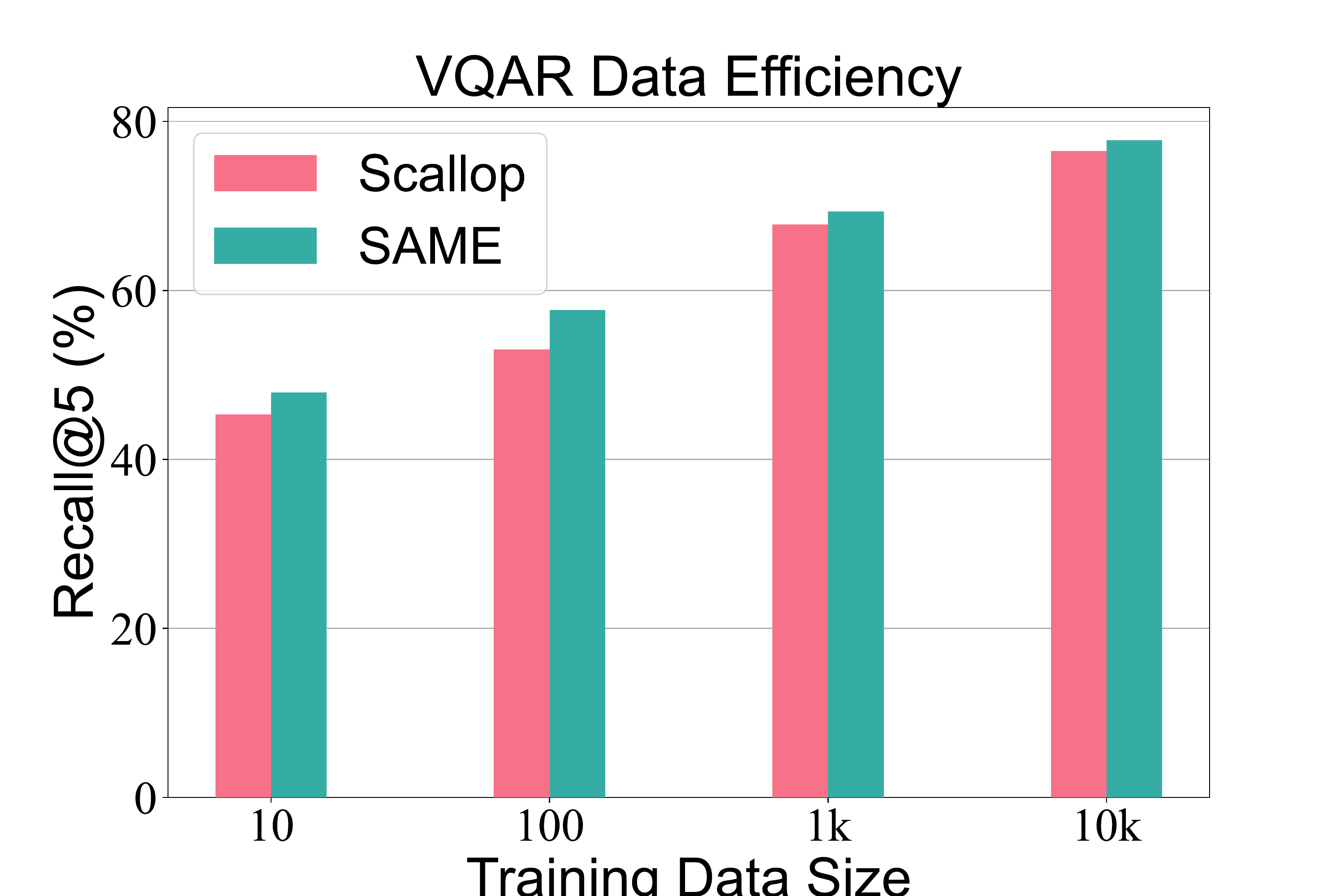}
        \caption{Data efficiency of SLASH with SAME and Scallop on different dataset sizes.}
        \label{fig:vqa-data-efficiency}
    \end{subfigure}
        \begin{subfigure}[b]{0.53\textwidth}
            \resizebox{\columnwidth}{!}{%
                \begin{tabular}{lrrrr}
                \toprule
                Train $\rightarrow$ & \multicolumn{2}{c}{10k $C_2$}  & \multicolumn{2}{c}{50k $C_{all}$} \\
    
                Test $\downarrow$  & \hspace{-0.3cm}Scallop & SAME   & Scallop & SAME \\
                    \midrule
                1k $C_2$ & 76.50& 77.76 & 84.34 & 79.26\\
                1k $C_3$ & 76.48& 73.75 & 81.97 & 72.95 \\
                1k $C_4$ & 77.00& 67.41 & 80.90 & 69.50\\
                1k $C_5$ & 79.30& 68.87 & 83.76 & 69.33\\
                1k $C_6$ & 76.98& 64.86 & 82.61 & 63.18 \\
            5k C$_{all}$ & 77.25 & 70.52 & 82.72 & 70.90 \\
                \bottomrule
            \end{tabular}
            }
        \caption{Performance of SLASH with SAME and Scallop on different clauses lengths.}
        \label{tab:vqa-performance}
    \end{subfigure}
    \label{fig:vqa-results}
    \caption{\textbf{Performance of SLASH on VQAR:}
    results on the data efficiency test (a) and generalization test for different clause lengths (b) trained on C\textsubscript{2} (left) and C\textsubscript{all} (right).}
\end{figure}

\textbf{Results -- } Fig.~\ref{fig:vqa-data-efficiency} presents insights on data efficiency: the recall@5 of test queries after training with 10, 100, 1k and 10k training samples on C\textsubscript{2}.
We see that SAME achieves greater data efficiency than Scallop due to the flexible number of potential solutions.

In Tab.~\ref{tab:vqa-performance}, the recall values are displayed for varying clause lengths to demonstrate our approach's generalizability and overall performance. The left side shows results for training on 10k samples on C\textsubscript{2} and the right side on C\textsubscript{all}.
SAME performs similarly to Scallop \cite{Scallop} on C\textsubscript{2} for both settings.
As the solution space grows exponentially with the complexity of the questions, we observe that the performance of SAME decreases compared with C\textsubscript{2} on more complex tasks.
Comparing Scallop's results with SAME, i.e. top-99\%, \citeA{Scallop} use top-10 for each programmatic query.
Scallop features directly weighted rules, while SLASH would have to emulate such rules.

In summary, the experimental results show that \emph{SLASH scales with SAME to VQA}.
Next, we study the scalability achieved by SAME as an ablation study.

\subsection{Scalability of SLASH}
\label{mnist-exp}

\begin{table}[t]
        \centering
        \begin{tabular}{lrrrrrrr}
        \toprule
        Neural &Task & SLASH &\multicolumn{4}{c}{top-k} & SAME\\
        Model & &  &k=1 & k=3 & k=5 & k=10 &  \\
        \midrule
        DNN & T1 &\gradient{98.80}&\gradient{98.68}&\gradient{98.81}& \gradient{98.60}&\gradient{98.69}&\gradient{98.56}\\
            & T2 &\gradient{98.85}&\gradient{98.81}&\gradient{98.76}& \gradient{98.68}&\gradient{98.68}&\gradient{98.82}\\
            & T3 &\gradient{98.75}&\gradient{98.77}&\gradient{98.77}& \gradient{98.74}&\gradient{98.77}&\gradient{98.71}\\
        \midrule
        PC  & T1 &\gradient{95.29}&\gradient{74.89}&\gradient{70.25}& \gradient{79.89}&\gradient{87.59}&\gradient{95.19}\\
            &T2 &\gradient{95.26}&\gradient{70.12}&\gradient{64.23}& \gradient{64.42}&\gradient{71.11}&\gradient{94.99}\\
            &T3 &\gradient{95.11}&\gradient{30.55}&\gradient{41.03}& \gradient{31.79}&\gradient{32.96}&\gradient{94.94}\\
        \bottomrule
        \end{tabular}
    \caption{\textbf{Regardless of how complex the task is, it is harder to choose the correct k in top-k than for top-k\% for PCs:}
    Accuracy Comparison between top-k and SLASH with and without SAME. We compare the method on three different Tasks, T1-T3:  $\mbox{\it sum2}$(\img{graphs/handwritten_digit_three.pdf}, \img{graphs/handwritten_digit_seven.pdf}, 10)  $\mbox{\it sum3}$(\img{graphs/handwritten_digit_three.pdf},\img{graphs/handwritten_digit_seven.pdf},\img{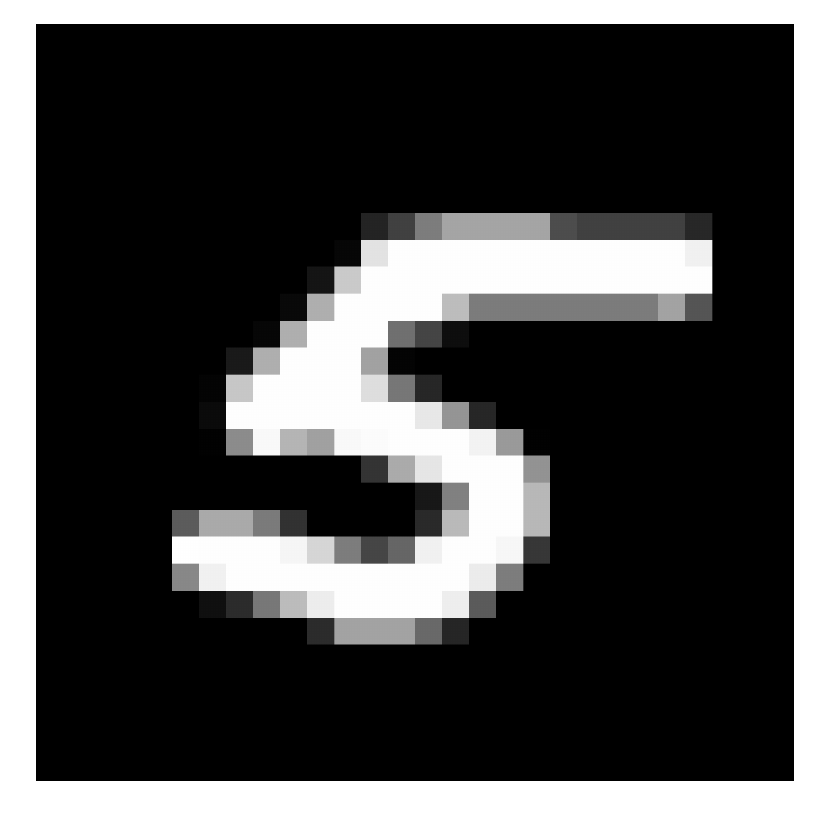},15) $\mbox{\it sum4}$(\img{graphs/handwritten_digit_three.pdf},\img{graphs/handwritten_digit_seven.pdf},\img{graphs/handwritten_digit_five.pdf},\img{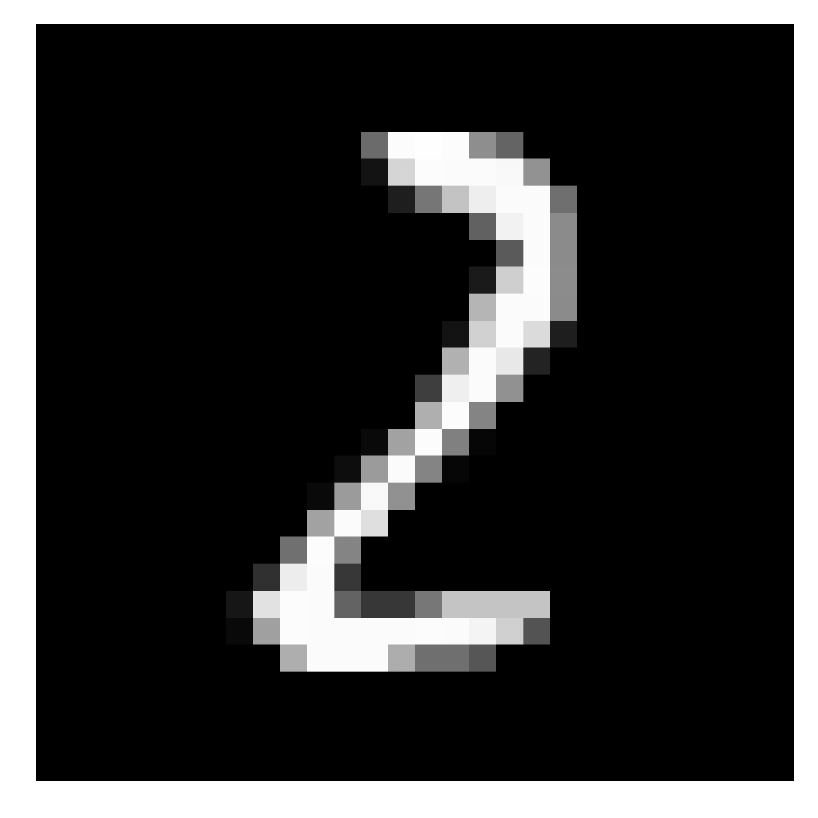},17).
    }
    \label{tab:mnist-partial-solution}
\end{table}

Inspired by \citeA{Scallop}, we explore how using different subsets of all potential solutions affects the performance and scalability of SLASH on the MNIST addition task. 

In the task of MNIST-addition \cite{DeepProbLog}, the goal is to predict the sum of two images from the MNIST dataset \cite{MNIST}, presented only as raw images.
During test time, however, a model should classify the images directly. 
Thus, the model does not receive explicit information about the depicted digits and must learn to identify digits via indirect feedback on the sum prediction. 
Using more than two images makes the task significantly harder, as an exponentially growing number of digit combinations has to be considered. 
Similar to the setup of Scallop \cite{Scallop}, we test on three different difficulty levels to evaluate the model's scaling capabilities. 
The difficulty ranges from task T1 with two images $\mbox{\it sum2}$(\img{graphs/handwritten_digit_three.pdf},\img{graphs/handwritten_digit_seven.pdf},10), to task T3 with four images $\mbox{\it sum4}$(\img{graphs/handwritten_digit_three.pdf},\img{graphs/handwritten_digit_seven.pdf},\img{graphs/handwritten_digit_five.pdf},\img{graphs/handwritten_digit_two.pdf},17). 
We use a probabilistic circuit (PC) and a deep neural network (DNN) as NPP for the same settings. 

The DNN used is the LeNet5 model \cite{LeNet}. 
When using the PC as NPP, we have extracted conditional class probabilities $P(C|X)$ by marginalizing the class variables $C$ to acquire the normalization constant $P(X)$ from the joint $P(X,C)$, and calculating $P(X|C)$.
The models using the NN architecture converge after one or two epochs and only get minor improvements in accuracy thereafter. 
For the PC architecture, the convergence takes more epochs and increases with the task difficulty.  
We report test accuracies after 10 and 20 epochs for the DNN and PC architecture, respectively. 
Tab.~\ref{tab:mnist-same-times} in App.~\ref{app:pc-convergence} shows the convergence of SLASH with PC as NPP on different tasks.

\textbf{Performance using subsets of all potential solutions -- } First, let us look at what happens if we prune away some potential solutions given our NPP probabilities. 
We compute the potential solutions in three ways: SLASH with all potential solutions, SLASH with a top-k variant (SLASH-top-k) and SLASH with SAME.
For top-k, we use CLINGO's minimization constraints to put the NPP output probabilities in the logic program, cf. App.~\ref{app:slash-programs}. 
The solver then gives us the potential solutions sorted by their probability $P_{\Pi(\bm{\theta})}(I)$, from which we keep the k most probable solutions.
For an example program for SLASH top-k, see App.~\ref{app:slash-programs}. 

\begin{table}[t]
    \resizebox{\columnwidth}{!}{%
        \begin{tabular}{lrrr|rrr}
        \toprule
         & \multicolumn{3}{c}{Accuracy after last Epoch} & \multicolumn{3}{c}{Average Time per Epoch} \\
         Method & \multicolumn{1}{c}{T1}& \multicolumn{1}{c}{T2}& \multicolumn{1}{c}{T3$^{*}$}   & \multicolumn{1}{c}{T1}& \multicolumn{1}{c}{T2}& \multicolumn{1}{c}{T3$^{*}$}\\
         \midrule
         Scallop top-10 &\gradient{98.95}&\gradient{99.12}&\gradient{97.47}& \ctime{0}{54}{10} & \ctime{5}{39}{53}& \ctime{21}{10}{0}\\
         DeepProbLog &\gradient{98.50}&\gradient{98.75}&\gradient{98.23}&\ctime{0}{8}{3}&\ctime{0}{15}{36}&\ctime{0}{34}{54}\\
         DeepStochLog &\gradient{96.96}&\gradient{97.49}&\gradient{97.54}&\ctime{0}{1}{23}&\ctime{0}{5}{49}&\ctime{0}{44}{27}\\
         NeurASP &\gradient{98.05}&\gradient{98.42}&\gradient{98.03}&\ctime{0}{3}{13}&\ctime{0}{32}{26}&\ctime{15}{28}{51}\\
         \midrule
         SLASH-DNN &\gradient{98.80}&\gradient{98.85}&\gradient{98.75}&\ctime{0}{0}{24}&\ctime{0}{1}{42}&\ctime{0}{51}{49}\\
         SLASH-PC &\gradient{95.29}&\gradient{95.26}&\gradient{95.11}&\ctime{0}{1}{9}&\ctime{0}{2}{27}&\ctime{0}{52}{22}\\
         SLASH-DNN top-10&\gradient{98.69}&\gradient{98.68}&\gradient{98.77}&\ctime{0}{0}{25}&\ctime{0}{1}{3}&\ctime{0}{25}{2}\\
         SLASH-PC top-10&\gradient{87.59}&\gradient{71.11}&\gradient{32.96}&\ctime{0}{1}{8}&\ctime{0}{1}{47}&\ctime{0}{26}{52}\\
         \midrule
         SAME-DNN &\gradient{98.56}&\gradient{98.82}&\gradient{98.71}&\ctime{0}{0}{17}&\ctime{0}{0}{17}&\ctime{0}{1}{35}\\
         SAME-PC &\gradient{95.19}&\gradient{94.99}&\gradient{94.94}&\ctime{0}{1}{3}&\ctime{0}{1}{23}&\ctime{0}{16}{40}\\
         \bottomrule    
        \end{tabular}
    }
    \caption{\textbf{SAME scales well with growing task complexity:} Test accuracy in \% and runtime comparison. The runtime is averaged over ten epochs for all methods. Light green indicates high accuracy or low time, while blue stands for the opposite. $(^{*})$ Please note, that since training Scallop and NeurASP would have taken too long, they were stopped after one epoch and therefore did not converge as the other DPPLs.}
    \label{tab:mnist-comparison}
\end{table}

Tab.~\ref{tab:mnist-partial-solution} lists the results for the test on partial solutions. 
SLASH and SAME achieve almost identical or slightly worse performance on all tasks and different NPPs.
With neural networks as our NPP, SLASH-top-k achieves similar performance for all k's compared to SLASH.
Using PCs as NPP, we get a worse performance. 
With increasing task difficulty, we lose most of the predictive performance of our model. 
With a high k on T1, most potential solutions are still covered, resulting in only a small drop in accuracy. For example, on T1 there are nine ways to add two digits to ten, which is the query with the most potential solutions.
With increasing task difficulty, though, many more potential solutions are not covered when selecting k=10 as in Scallop~\cite{Scallop} since there are 73 for T2 and 633 for T3.
At the beginning of training, our model gives us uniform predictions over all digits, as it has not learned anything yet. 
Therefore, the randomness of model initialization influences which solution falls into the top-k range.
If we prune the true solution, our model cannot learn to detect the correct class with that query, and it has to rely on other queries that might have the true solution in the top-k range. 
Empirically, we see that with DNNs, we can still learn to detect digits, while with PCs, we cannot. 
We argue that the DNN architecture is more robust to these incorrect inputs and, over time, accumulates an increasing proportion of the correct digits in the top-k selection because it is better suited for object detection equipped with the visual inductive biases of convolutional layers. 
PCs, on the other hand, learn false classes at the beginning and reinforce the false prediction by repeatedly predicting them as most likely.

On the contrary, SAME works on both PCs and DNNs as it only prunes certainly unlikely options.
At first, we do not prune anything, and over time, after learning, we can safely regard the unlikely solutions, which explains why SAME is the better choice for both NNs and PCs.

\textbf{SAME reduces training time by pruning unlikely outcomes -- } 
After seeing that SLASH with SAME achieves on-par performance, we now want to look at the time savings we get by using it. 
Tab.~\ref{tab:mnist-comparison} shows the average training time per epoch and the test accuracy. 
We provide results for other state-of-the-art DPPLs: Scallop \cite{Scallop}, DeepProbLog \cite{DeepProbLog}, its cousin DeepStochLog \cite{DeepStochLog}, and NeurASP \cite{NeurASP}.
These DPPLs again use the LeNet5 architecture \cite{LeNet}.
For Scallop and NeurASP, we report the accuracy after one epoch on T3, as the training time for ten epochs would take almost a week.

\begin{table}[]
    \centering
    \begin{tabular}{lrrrr}
    \toprule
         Model  & Epochs & T1 & T2 & T3\\
         \midrule
         SAME-DNN & 1   &\ctime{0}{0}{21} &\ctime{0}{0}{39}&\ctime{0}{13}{37}\\
                   & 1-10 &\ctime{0}{0}{17} &\ctime{0}{0}{17}&\ctime{0}{1}{35}\\
                   & 2-10 &\ctime{0}{0}{17} &\ctime{0}{0}{15}&\ctime{0}{0}{14}\\

        \midrule
         SAME-PC & 1    &\ctime{0}{1}{14}&\ctime{0}{2}{32}&\ctime{0}{53}{26}\\
                  & 1-20 &\ctime{0}{1}{3} &\ctime{0}{1}{23}&\ctime{0}{16}{40}\\
                  & 2-20 &\ctime{0}{1}{2} &\ctime{0}{1}{15}&\ctime{0}{12}{35}\\

         \bottomrule
    \end{tabular}
    \caption{\textbf{Due to pruning, SAME gets faster in later iterations:} 
    The average time per epoch is shown for different epochs.
    }
    \label{tab:mnist-same-times}
\end{table}
 
SLASH with and without SAME achieves state-of-the-art accuracy similar to the other models on all task difficulties using the same DNN architecture. 
We further observe that the test accuracy of SLASH with a PC NPP is slightly below the other DPPLs. 
However, this may be since a PC, compared to a DNN, is learning a true mixture density rather than just conditional probabilities. 
Note that optimal architecture search for PCs, e.g., for computer vision, is an open research question.

Regarding training time, we see that top-k yields small improvements. 
With SAME we improve the training time by a huge fraction when considering a large number of potential solutions. 
For example, on T3 with NNs, we only need 3\% of SLASH's original training time over 10 epochs (see Fig.~\ref{fig:same-savings}). 
Tab.~\ref{tab:mnist-same-times} gives a more detailed overview of SLASH training times with SAME. 
Interestingly, after one epoch of training, the average runtime per epoch for epochs 2-10 is the same for all three difficulties for the DNN, as the model converges for the most part after the first epoch. 
It is even a bit faster on T3 because the number of queries is less on the T3 dataset (60k samples/number of images per query).

These evaluations, in summary, show that SAME is an efficient extension of SLASH which saves a lot of computing resources while yielding tiny to no differences in performance.

\subsection{Object-centric learning}
\label{sec:object-centric-learning}
Now, we turn to a very different task of object-centric set prediction. 
We presume that recent advancements in object-centric learning can be further improved by integrating such neural components into DPPLs and adding logical constraints about objects and their properties \cite{GreffIODINE,LinSPACE,SlotAttention}. 
In similar manner, we want to find out how much SAME speeds up SLASH possibly without loss of performance.

\begin{figure}[t]
\begin{subfigure}{.5\textwidth}
    \centering
    \includegraphics[scale=0.22]{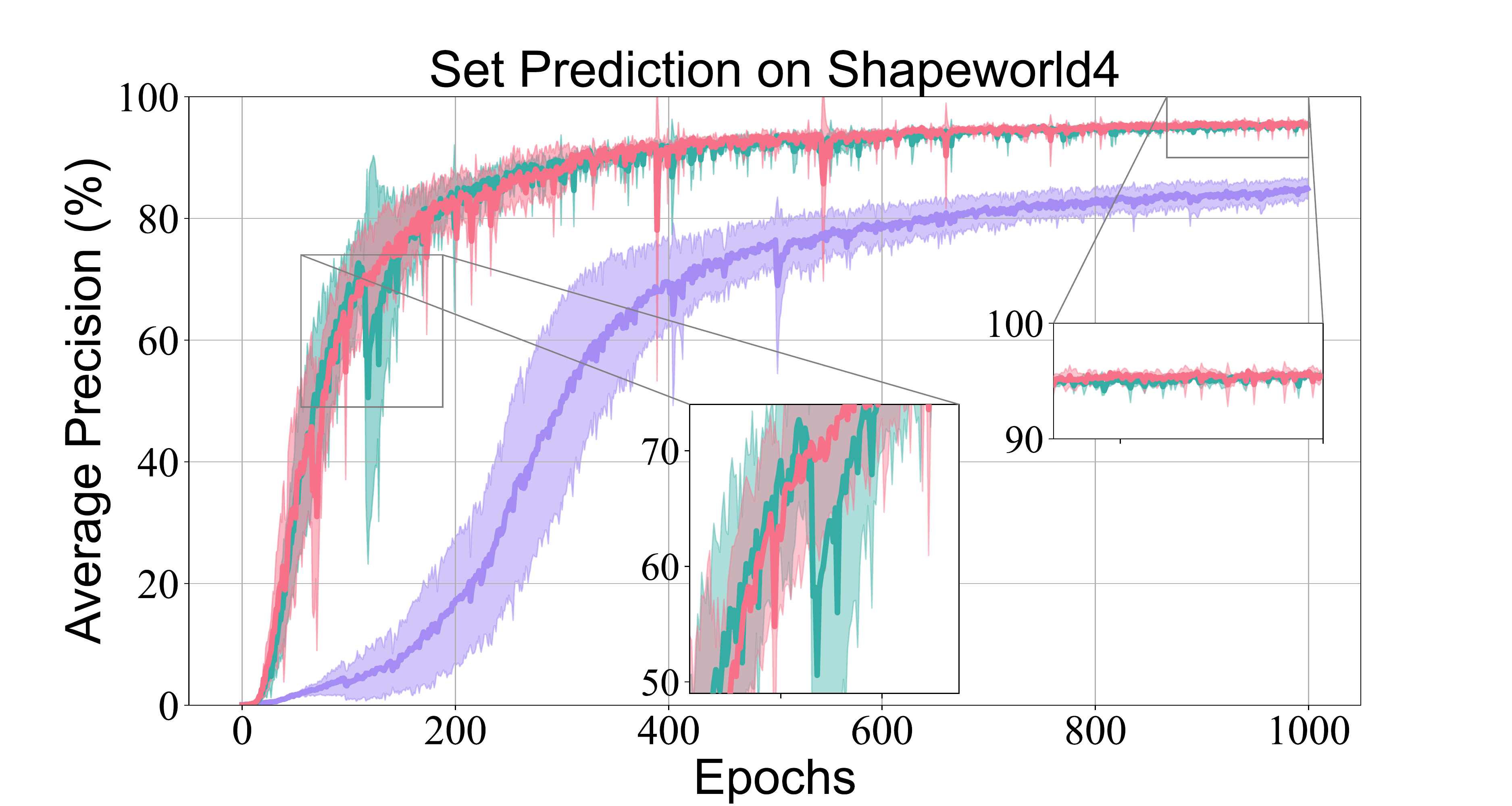}
  \label{fig:ap2}
\end{subfigure}%
\begin{subfigure}{.5\textwidth}
   \centering
    \includegraphics[scale=0.22]{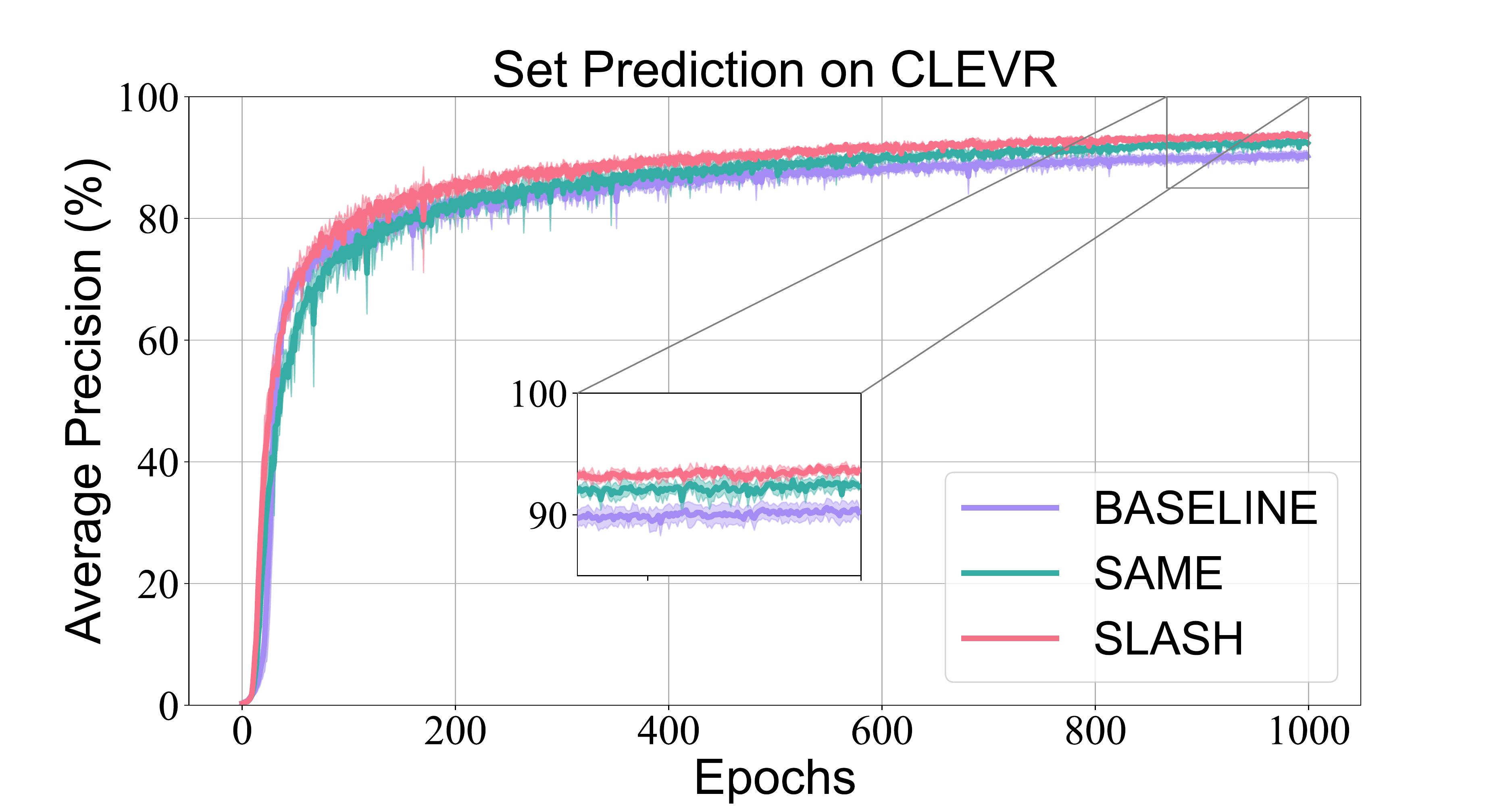}
  \label{fig:ap1}
\end{subfigure}
\vspace{-0.2in}
\caption{\textbf{SLASH can converge faster:} Average Precision on ShapeWorld4 (left) and CLEVR (right).
SLASH converges faster on ShapeWorld4 compared to the Baseline. During training, we observe temporary crashes in AP, which get smaller over time (see zoomed windows).
Furthermore, the standard deviation is much smaller for SLASH than for the baseline. 
On CLEVR, all three models converge similarly after roughly 200 epochs.
SLASH performs slightly better than SAME, which, in turn, performs a little better than the Baseline.}
\label{fig:ap}
\end{figure}

For set prediction, a model is trained to predict the discrete attributes of a set of objects in an image (\textit{cf.} Fig.~\ref{fig:npp} in the top-left corner for an example CLEVR image). 
The difficulty lies therein that the model must match an unordered set of corresponding attributes of various objects with its internal representations of the image.

The slot attention module introduced by \citeA{SlotAttention} allows for an attractive object-centric approach to this task. 
Specifically, this module represents a pluggable, differentiable module that can be easily added to any architecture. 
Through a competitive softmax-based attention mechanism, the model can enforce the binding of specific parts of a latent representation into permutation-invariant, task-specific vectors called slots.

We train SLASH with and without SAME based on NPPs consisting of a shared slot encoder and separate PCs, each modelling the mixture of latent slot variables and the attributes of one category, e.g., color. 
For each dataset, ShapeWorld4 and CLEVR, we have four NPPs in total. 
Finally, the model is trained via queries exemplified in Fig.~\ref{fig:program_shapeworld_query} in App. 
We refer to this configuration as \emph{SLASH Attention}.

We compare SLASH Attention to a baseline of slot attention encoder using single multicategorical MLP and Hungarian loss to predict object properties from the slot encodings as in \cite{SlotAttention}.
The key difference between these two models lies in the employed logical constraints in SLASH Attention. In their work, \citeA{SlotAttention} utilize a single MLP trained via Hungarian loss, i.e., they assume shared parameters for all attributes.
In comparison, in SLASH attention, we make an independence assumption about the parameters for the object attributes and encode this via logical constraints.
We refer to App.~\ref{app:slash-programs} for the program.

One limitation is that matching objects to slots has $n!$ possible assignments. 
To overcome this problem, we adopt a similar strategy to external functions in CLINGO. 
We use Hungarian matching \cite{hungarian} and make the resulting assignments a part of the logic program. 
This allows us to scale polynomial with the number of objects and enables SLASH for training on CLEVR, which can contain up to ten objects per image. 
The results of these experiments can be found in Tab.~\ref{fig:slash_attention}.

On ShapeWorld4, we observe that the average precision after convergence on the held-out test set with SLASH Attention is greatly improved to that of the baseline model.
More interesting, SAME provides the best results in this setting while having the smallest deviation, cf. Fig.~\ref{fig:ap} (left).
ShapeWorld4 has significantly fewer data entries than CLEVR, which may account for the huge improvement in performance with SLASH Attention. 
This is evidence that we are moving closer to knowledge-rich AI.
Additionally, we observe that SLASH Attention reaches the average precision value of the baseline model in much fewer epochs.
On CLEVR, this tendency also holds, but the difference in performance is smaller, but we still get around 2-3\% more average precision with SLASH and SAME.

Regarding the training times, we observed that in the case of ShapeWorld4, using SAME allows truncating the training window by \textbf{44.47\%}, compared to the results of SLASH without SAME.
For CLEVR, we not only obtained a solution, but also are getting it \textbf{21.4\%} faster thanks to SAME.


\begin{table}[t]
        \centering
        \begin{tabular}{lcc|cc}
            \toprule
            & \multicolumn{2}{c}{Accuracy} & \multicolumn{2}{c}{Time} \\
            Method & \multicolumn{1}{c}{ShapeWorld4} & \multicolumn{1}{c}{CLEVR}   & \multicolumn{1}{c}{ShapeWorld4}& \multicolumn{1}{c}{CLEVR}\\
            \midrule
            Slot Att. & \gradient{85.25} & \gradient{90.21} & \dtime{0}{2}{36}{0} & \dtime{1}{2}{26}{0}\\
            SLASH & \gradient{94.98} & \gradient{93.47} & \dtime{1}{3}{53}{0} & \dtime{6}{16}{49}{0}\\
            SAME & \gradient{95.21} & \gradient{92.50} & \dtime{0}{15}{29}{0} & \dtime{5}{6}{24}{0}\\
            \bottomrule
        \end{tabular}
        \caption{\textbf{SLASH can improve Slot Attention:} Test average precision and training times for the Slot Attention baseline and SLASH with and without SAME.}
        \label{fig:slash_attention}
\end{table}

These observations suggest that SAME applies to any form of NPPs and is a good step towards unraveling the solving bottleneck to lift the symbolic overhead.
Nonetheless, there is a difference in the number of learnable parameters between the neural baseline and SLASH attention.
Namely, SLASH attention consists of four PCs, for which the time spent on forward- and backward-pass is higher compared to the single multicategorical DNN used in the slot attention module.
We refer to App.~\ref{app:nesy-train-times} for the in-detail discussion.
Finally, we draw attention to the fact that the symbolic overhead is a direct result of all DPPLs under consideration using the CPU for high-level reasoning, while low-level perception is based on the GPU. 
To further reduce the symbolic overhead, tight integration of solving with neural processing may be a promising research direction.

\textbf{Summary of Empirical Results.} 
All empirical results together demonstrate that the expressiveness and flexibility of SLASH are highly beneficial and improve upon the state-of-the-art: one can freely combine what is required to solve the underlying task --- (deep) neural networks, PCs, and logic.
The experiments demonstrate -- SAME to be the inherent extension of SLASH.
Further, the results indicate that utilization of SAME comes with a tiny, if any, performance loss comparatively with the analytical weighted model counting.

\section{Related Work}
\textbf{Neuro-Symbolic AI} can be divided into two lines of research, depending on the starting point, though both have the same final goal: to combine low-level perception with logical constraints and reasoning.
A key motivation of Neuro-Symbolic AI \cite{GarcezLG2009,MaoNSCL,HudsonM19,GarcezGLSST19,JiangA20,GarcezLamb} is to combine the advantages of symbolic and neural representations into a joint system. 
This is often done in a hybrid approach where a neural network acts as a perception module that interfaces with a symbolic reasoning system, e.g., \cite{MaoNSCL,yiNSVQA}.
The goal of such an approach is to mitigate the issues of one by the other, e.g., using the power of symbolic reasoning systems to handle the generalizability issues of neural networks and handle the difficulty of noisy data for symbolic systems via neural networks. 
Recent work has also shown the advantage of approaches for explaining and revising incorrect decisions \cite{ciravegna2020human,stammerNeSyXIL}. 
However, many of these previous works train the sub-symbolic and symbolic modules separately.

\textbf{Deep Probabilistic Programming Languages (DPPLs)} are programming languages that combine deep neural networks with probabilistic models and allow a user to express a probabilistic model via a logic program. Similar to neuro-symbolic architectures, DPPLs thereby unite the advantages of different paradigms. DPPLs are related to earlier works such as Markov Logic Networks (MLNs) \cite{MLN}. Thereby, the binding link is the Weighted Model Counting (WMC) introduced in LP\textsuperscript{MLN} \cite{LPMLN}. 
Several DPPLs have been proposed by now, among which are Pyro \cite{Pyro}, Edward \cite{Edward}, DeepProbLog \cite{DeepProbLog}, DeepStochLog \cite{DeepStochLog}, NeurASP~\cite{NeurASP}, and Scallop~\cite{Scallop}.

To resolve the scalability issues of DeepProbLog, which uses Sentential Decision Diagrams (SDDs) \cite{SDD} as the underlying data structure to evaluate queries, NeurASP \cite{NeurASP}, offers a solution by utilizing ASP~\cite{dimopoulos1997encoding,soininen1999developing,marek1999stable,ASP-Core-2}. 
In contrast to query evaluation in Prolog \cite{colmerauer1996birth,Prolog}, which may lead to an infinite loop, many modern answer set solvers use Conflict-Driven-Clause-Learning (CDPL), which, in principle, always terminates. 
In this way, NeurASP changes the paradigm from query evaluation to model generation, i.e., instead of constructing an SDD or a similar knowledge representation system, NeurASP generates a set of all potential solutions (one model per solution) and estimates the probability for the truth value of each of these solutions. Of those DPPLs that handle learning in a relational, probabilistic setting and in an end-to-end fashion, all of these are limited to estimating only conditional class probabilities.

Another research branch focuses on approximate inference for DPPLs to allow scaling to harder problems. The goal is to incorporate probabilities into the solving process to obtain only a subset of all proofs. 
\citeA{ApproxDeepProbLog} propose an A*-like search for proofs, and \citeA{Scallop} introduce a top-k mechanism based on Datalog to only keep likely proofs.
In ASP, a program is first grounded and then solved, sometimes making the grounding itself a bottleneck. 
Existing work, therefore, aims at grounding on demand. The two main candidates are Lazy Grounding \cite{LazyGrounding} or Magic Sets for ASP \cite{magic-sets-asp}. 
To the best of our knowledge, both techniques have not been applied in a probabilistic setting with ASP yet.

\textbf{Visual Question Answering} has seen a lot of attention from the computer vision and natural language processing community. We refer to \cite{VQA-Review} and \cite{VQA-Review-new} for a detailed review. 
Recently, more neuro-symbolic approaches to VQA have been proposed. \citeA{yiNSVQA} proposed a model which creates a structural scene representation of the image, parses a natural language question into a program, and then executes the program to obtain an answer. 
A few works utilize logic programming: Scallop's \cite{Scallop} top-k approach allows for answering complex reasoning questions on real-world images. \citeA{vienne-nesy} showed how ASP could be used on top of the outputs of a pretrained YOLO network to answer CLEVR questions~\cite{CLEVR}.

\section{Conclusions}
We introduce SLASH, a novel DPPL that integrates neural computations with tractable probability estimates and logical statements. 
The key ingredient of SLASH to achieve this are Neural-Probabilistic Predicates (NPPs) that can be flexibly constructed out of neural and/or probabilistic circuit modules based on the data and underlying task. 
With these NPPs, one can produce task-specific probability estimates. 
The details and additional prior knowledge of a task are neatly encompassed within a SLASH program with only a few lines of code. 
Finally, via ASP and Weighted Model Counting, the logic program and probability estimates from the NPPs are combined within SLASH to estimate the truth value of a task-specific query. 
Additionally, the SAME technique addresses the question of scalability. 
Proven to converge to only one solution, SAME is the inherent extension of SLASH generally applicable to any problem.

Our experiments on the VQAR dataset show the power, efficiency, and scalability of SLASH, paving the way to handle extremely difficult real-world applications. 
As one of many consequences, we found the following shortcomings to be resolved in future work.
First, VQAR admits bigger parts of a program are optional to answer the programmatic query and, thus, should be ignored during grounding.
Consequentially, we need ``grounding on demand''. 
SAME is a form of stochastic lazy grounding, and thus is evident to help reduce the computation costs for NPP. 
Thus, it remains to be seen if and how similar technique(s) can be used for grounding complete programs.
Second, should there be an exponential number of potential solutions, as in some VQAR queries, we cannot answer the query anymore. 
Weighted rules and facts might be insightful in finding ways to navigate solution spaces more efficiently.
Finally, for WMC to be computed most efficiently regardless of the number of potential solutions and the queries, it must take place simultaneously with solving, i.e., becoming an inseparable part of it.

Apart from that, our ablation study provided detailed insights on the computation speed of SAME, improving upon previous DPPLs in the benchmark MNIST-Addition tasks yet retaining the performance. 
Additionally, invoking Python routines allowed for the seamless invocation of the Hungarian matching algorithm into SLASH Attention.
Together with SAME, we solved the task of object-centric set prediction for the CLEVR dataset, which none of the previous DPPLs has tried to solve yet, and reduced the training time of SLASH. 


With SLASH on the set prediction task, we effectively use elements of functional programming within SLASH.
Similar, the used ASP-solver CLINGO can invoke Python routines at the grounding time via external functions.
These pave the way for merging functional programming with SLASH.
Neural Logic Machines \cite{nlm} serve as an example of a similar combination.
Going in this direction will allow us to treat logically constrained regression problems, which would benefit fundamental sciences such as particle physics.
\citeA{yu2021uai_momogps} show how PCs can be used for multi-output regression tasks, and it appears to be the natural next step to integrate them in SLASH.

\acks{This work was partly supported by the Federal Minister of Education and Research (BMBF) and the Hessian Ministry of Science and the Arts (HMWK) within the National Research Center for Applied Cybersecurity ATHENE,  the ICT-48 Network of AI Research Excellence Center ``TAILOR'' (EU Horizon 2020, GA No 952215, and the Collaboration Lab with Nexplore ``AI in Construction'' (AICO). It also benefited from the BMBF AI lighthouse project, the Hessian research priority programme LOEWE within the project WhiteBox, the HMWK cluster projects ``The Third Wave of AI'' and ``The Adaptive Mind'', the German Center for Artificial Intelligence (DFKI) project ``SAINT''.}


\newpage

\appendix

\section{SLASH Programs}
\label{app:slash-programs}
Here, the interested reader will find the SLASH programs which we compiled for our ablation studies on MNIST addition and Object-centric learning.
 Fig.~\ref{fig:program_shapeworld} and Fig.~\ref{fig:program_shapeworld_query} -- for the set prediction task with slot attention encoder.
\begin{figure}[ht!]
    \centering
    {
\begin{lstlisting}[language=Python]
# Define images
img(i1). img(i2).
# Define Neural-Probabilistic Predicate
npp(digit(X), [0,1,2,3,4,5,6,7,8,9]) :- img(X).
# Define the addition of digits given two images and the resulting sum
addition(A, B, N) :- digit(+A, -N1), digit(+B, -N2), N = N1 + N2.
\end{lstlisting}}
    \caption{SLASH Program for MNIST addition with two images.} 
    \label{fig:program_mnist_addition}
\end{figure}
\begin{figure}[ht!]
    \centering
    {
\begin{lstlisting}[language=Python]
# Is 7 the sum of the digits in img1 and img2?
:- addition(i1, i2, 7)
\end{lstlisting}}
    \caption{Example SLASH Query for MNIST addition.}
    \label{fig:program_mnist_addition_query}
\end{figure}
\\The following code snippet shows the weak constraints for SLASH top-k.
\begin{figure}[ht!]
    \centering
    {
\begin{lstlisting}[language=Python]
# weak constraints for image 1
:~ digit(1,i1,0). [1866, 0, 0]
:~ digit(1,i1,1). [2901, 0, 1]
... 
:~ digit(1,i1,9). [2468, 0, 9]
# weak constraints for image 2
:~ digit(1,i2,0). [1761, 1, 0]
:~ digit(1,i2,1). [2922, 1, 1]
...
:~ digit(1,i2,9). [2517, 1, 9]
\end{lstlisting}}
    \caption{Weak constraints for SLASH top-k }
    \label{fig:program_mnist_addition_weak_constraint}
\end{figure}
\\In the brackets, the first value is the probability of the corresponding ground atom in log space. 
The second and third values together make up a unique identifier for the belonging atom, which is used by CLINGO.

\begin{figure}[ht!]
    {
\begin{lstlisting}[language=Python]
# Define slots
slot(s1). slot(s2). slot(s3). slot(s4).
# Define identifiers for the objects in the image
# (there are up to four objects in one image).
obj(o1). obj(o2). obj(o3). obj(o4).
# Assign each slot to an object identifier
{assign_one_slot_to_one_object(X, O): slot(X)}=1 :- obj(O).
# Make sure the matching is one-to-one between slots
# and objects identifiers.
:- assign_one_slot_to_one_object(X1, O1),
   assign_one_slot_to_one_object(X2, O2),
   X1==X2, O1!=O2.
# Define all Neural-Probabilistic Predicates
npp(color_attr(X), [red, blue, green, grey, brown,
					   magenta, cyan, yellow, bg]) :- slot(X).
npp(shape_attr(X), [circle, triangle, square, bg]) :- slot(X).
npp(shade_attr(X), [bright, dark, bg]) :- slot(X).
npp(size_attr(X), [big,small,bg]) :- slot(X).
# Object O has the attributes C and S and H and Z if ...
has_attributes(O, C, S, H, Z) :- slot(X), obj(O),
								 assign_one_slot_to_one_object(X, O),
								 color(+X, -C), shape(+X, -S),
								 shade(+X, -H), size(+X, -Z).

\end{lstlisting}}
    \caption{SLASH Program for ShapeWorld4.}
    \label{fig:program_shapeworld}
\end{figure}

\begin{figure}[ht!]
    {
\begin{lstlisting}[language=Python]
# Does object o1 have the attributes red, circle, bright, small?
:- has_attributes(o1, red, circle, bright, small)
\end{lstlisting}}
    \caption{Example SLASH Query for ShapeWorld4 experiments. In other words, this query corresponds to asking SLASH: ``Is object 1 a small, bright red circle?''.}
    \label{fig:program_shapeworld_query}
\end{figure}

\newpage 

\section{VQA Program Encoding and dealing with complexity}
\label{app:experimental-details-vqa}
In this section, we will explain how the SLASH program for the VQA task is constructed and how we deal with the complexity of the task and thus avoid producing an infeasible number of potential solutions for difficult questions.
As depicted in Fig.~\ref{fig:vqa-overview}, the VQA task comprises multiple parts in the SLASH program. One thing to highlight here is the length of the program, which usually has more than 3k lines.

The KG makes 1424 ``is-a'' tuples and 1963 ``object-attribute-relation'' triplets, as well as six rules for the fixed part of every program. 
For $n$ objects, the SG includes $n$ attributes, $n$ names, and $n*(n-1)$ obj-to-obj relations, excluding relations of objects to themselves. 
Each object can have multiple attributes at once, so each attribute is modelled as a NPP with two outcomes: having or not having the attribute. 

Fig.~\ref{fig:vqar-example} shows two images, object bounding boxes, and a target rule specifying what targets should entail.
E.g., the provided target query from C\textsubscript{2} in Fig.~\ref{fig:vqar-traffic} is depicted.
\begin{lstlisting}[language=Python]
:- not target(0); not target(1).
:- target(2).:- target(3).:- target(4).:- target(5).:- target(6).
\end{lstlisting}\label{fig:program_vqa_query}

It restricts objects 0 and 1 to be targets, while others are not. 
Combined with the stated target rule, the name NPP outcomes of objects 0 and 1 are restricted to inferring a name that can be substituted for variable N0 in the oa\_rel(is\_used\_for, N0, controlling\_flows\_of\_traffic) predicate. 
From the knowledge graph, we can infer for N0 to be replaced by ``traffic lights''.
In this case, all other names for the non-target objects are restricted to not being traffic lights. 
They can take on all other 499 outcomes of the name NPP. 
Attributes and relations are not restricted as well by the query.
CLINGO's show statements are used to show exclusively the predicates of the programmatic query in any potential solution.
For the example under consideration, the show rule is depicted below.
\begin{lstlisting}[language=Python]
#show. 
#show name(O0, X) : target(O0), name(O0,X), name(O0,N0), oa_rel(is_used_for,N0,controlling_flows_of_traffic).
\end{lstlisting}\label{fig:program_vqa_show_statement}
These lines tell CLINGO to itemize name predicates of objects which satisfy all target predicates. 
Upon solving, we obtain the following potential solutions.
\begin{align*}
    &\text{\{target(0), name(0,traffic\_lights)\}}\\
    &\text{\{target(1), name(1,traffic\_lights)\}}
\end{align*}
Moving on to the more complex example of C\textsubscript{5} in Fig.~\ref{fig:vqar-sheep}. 
Here, our target rule consists of five predicates. 
Two name predicates restrict the target to be animals and objects. 
Additionally, two relation predicates specify the relation of the target object O2 to other objects. 
For such a query, computing all potential solutions can be infeasible. 
Particularly, we have 16 objects which form 16*(16-1)=240 relation NPPs. 
Substituting them into the target relations at once can quickly lead to millions of potential solutions, should a programmatic query contains multiple relation predicates.
Instead, we employ iterative solving. 
At each iteration, the next five relations with the highest probability are added until we have 100 potential solutions or a specified timeout of 30 seconds is reached.

In the last step, SAME helps out by pruning unlikely NPP outcomes. 
The target rule of this example stipulates only one name predicate should be an object.
From the KG, we see the ontological concept telling us about what falls under the category of objects, such as furniture, vehicles, or animals. 
Asking for these broader categories restricts the NPP outcomes only partially. 
In the case of objects, most Name NPP's outcomes are part of this category. 
Here again, for some queries, this can make computing all potential solutions infeasible. 
Our solution – combining top-k pruning with SAME: to keep the k most probable outcomes for each Name NPP and to prune more with SAME. 
Precisely, SAME will point to the Name outcome, that is, the actual one belonging to the object category and will prune any remaining ones.

\newpage 

\section{Experimental Details}
\label{app:experimental-details}

\subsection{PC Convergence}
\label{app:pc-convergence}
\begin{figure}[ht!]
    \centering
        \includegraphics[scale=0.21]{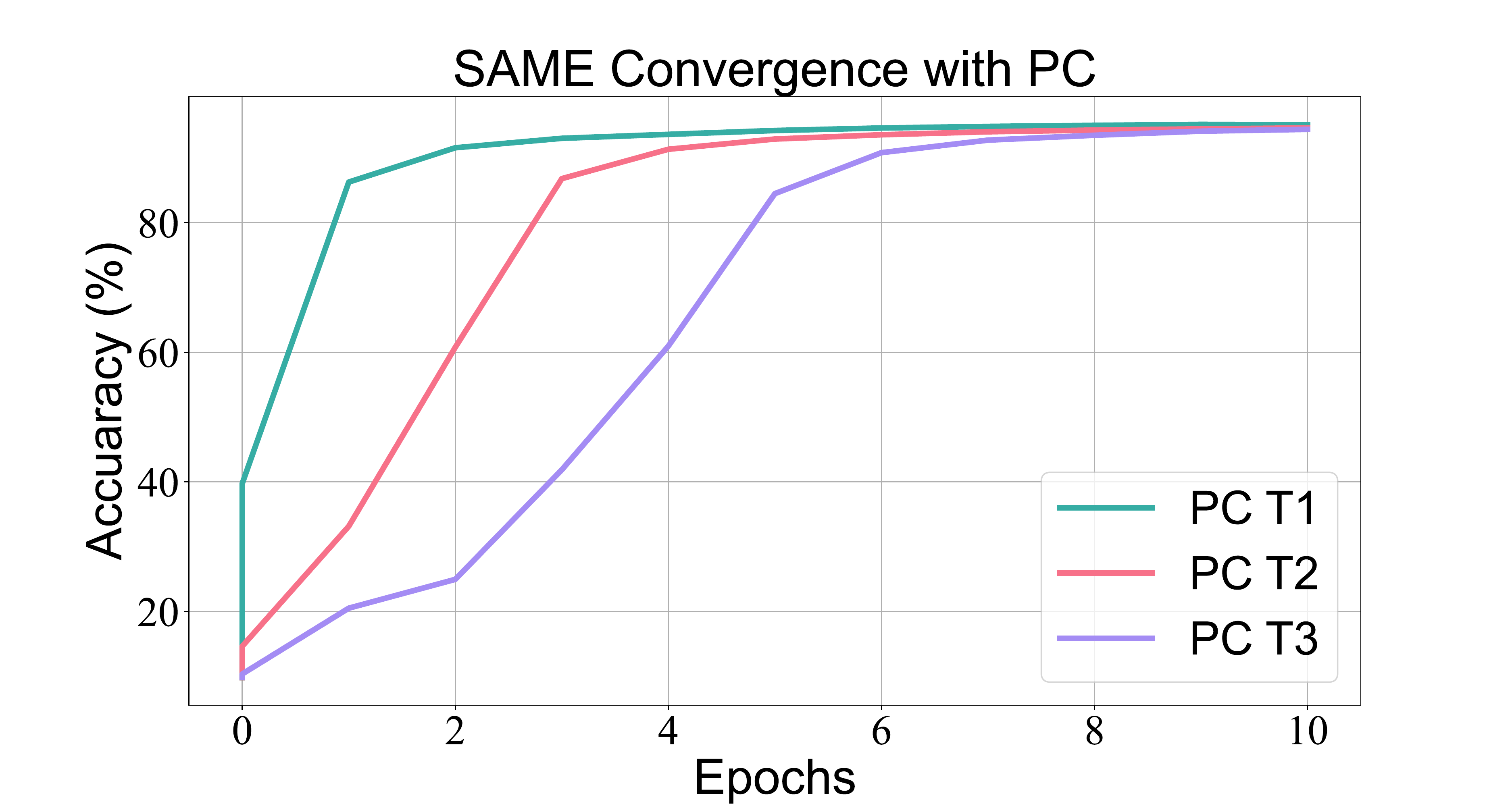}
    \caption{Convergence with SAME-PC on MNIST Addition}
    \label{fig:pc-convergence}
\end{figure}

Fig.~\ref{fig:pc-convergence} shows the convergence of SAME with PCs on the T1, T2 and T3. The accuracy converges to the almost same value, and it can be seen that the harder the task, the more epochs it takes to converge.

\subsection{ShapeWorld4 Generation}

The ShapeWorld4 dataset was generated using the original scripts of \cite{ShapeWorld}\footnote{https://github.com/AlexKuhnle/ShapeWorld}). The exact scripts will be added together with the SLASH source code.

\subsection{Average Precision computation}

For the baseline slot encoder experiments on the Object-centric learning tasks, we measured the average precision score as in \cite{SlotAttention}. 
In comparison to the baseline slot encoder, when applying SLASH Attention, however, we handled the case of a slot not containing an object, e.g., only background variables, differently. 
Whereas \citeA{SlotAttention} add another binary identifier to the multi-label ground truth vectors, we have added a background (bg) attribute to each category (\textit{cf.} Fig.~\ref{fig:program_shapeworld}). 
A slot is thus considered to be empty (i.e., not containing an object) if each NPP returns a high conditional probability for the \textit{bg} attribute.

\subsection{Model Details}
\label{app:model-details}
For those experiments using NPPs with PC, we have used Einsum Networks (EiNets) for implementing the probabilistic circuits. 
EiNets are a novel implementation design for SPNs introduced by \citeA{EiNets} that minimizes the issue of computational costs that initial SPNs had suffered. 
This is accomplished by combining several arithmetic operations via a single monolithic einsum-operation.

For all experiments, the ADAM optimizer \cite{ADAM} with $\beta_{1} = 0.9$ and $\beta_{2} = 0.999$, $\epsilon = 1e-8$ and no weight decay was used.

\textbf{VQA Experiments}
The architecture for the VQA experiments is the same as in \cite{Scallop} and is shown in Tab.~\ref{tab:vqa_architecture}.
The name, relation, and attribute classifier share the same architecture.
A YOLO network produces object features of size 2048 which are fed into the classifiers.
The relation classifier takes as input the features and bounding boxes of two objects, resulting in an input dimension of $4104 =(2048+4)*2$.
For the name and relation classifier, a Softmax is used.
The attribute classifier has a Sigmoid activation, encoding multiple attributes over each output neuron.

\begin{table}[ht!]
\centering
    \begin{tabular}{cccc}
        \toprule
        Type                 & Size/Channels & Activation & Comment \\
        \midrule
        MLP                  & input\_dim, 1024    & ReLU       & -\\
        BatchNorm + Dropout & 1024 & - & dropout-rate 0.3/0.5\\

        MLP*                  & 1024, 1024    & ReLU       &  \\
        BatchNorm + Dropout* & 1024 & - & dropout-rate 0.3 \\
        
        MLP                  & 1014, num\_classes    & Softmax/Sigmoid       & - \\
    
        \bottomrule
    \end{tabular}
\caption{VQA Neural Model. Layers marked with * are only used in the attribute and name classifier. \label{tab:vqa_architecture}}
\end{table}

\textbf{MNIST-Addition Experiments} For the MNIST-Addition experiments, we ran all baseline programs with their original configurations, as stated in \cite{Scallop,DeepProbLog,DeepStochLog,NeurASP}, respectively. For the MNIST Addition experiments, we have used the same neural module as in the baselines when training SLASH and SAME with the neural NPP represented in Tab.~\ref{tab:lenet-mnist}. 
When using a PC NPP, we have used an EiNet with the Poon-Domingos (PD) structure \cite{SPNs} and normal distribution for the leaves. 
The formal hyperparameters for the EiNet are depicted in Tab.~\ref{tab:einet_mnist}.
The learning rate and batch size for SLASH and the baselines are shown in Tab.~\ref{tab:hyperparams}.

\begin{table}[ht!]
\centering
    \begin{tabular}{ccc}
    \toprule
    Model & learning rate & batch size\\
    \midrule
    Scallop & 0.001  & 64 \\
    DeepProbLog & 0.0001 & 2 \\
    DeepStochlog & 0.001 & 100\\
    NeurASP  & 0.001  & - \\
    \midrule
    SLASH-DNN & 0.005 & 100 \\
    SLASH-PC & 0.01 & 100\\
    \bottomrule
    \end{tabular}
    \caption{Learning rate and batch size for the baselines and SLASH.}
    \label{tab:hyperparams}
\end{table}

\begin{table}[ht!]
\centering
    \begin{tabular}{cccc}
        \toprule
        Type                 & Size/Channels & Activation & Comment \\
        \midrule
        Encoder              & -             & -          & -       \\
        Conv 5 x 5           & 1x28x28       & -          & stride 1 \\
        MaxPool2d            & 6x24x24       & ReLU       & kernel size 2, stride 2 \\ 
        Conv 5 x 5           & 6x12x12       & -          & stride 1 \\ 
        MaxPool2d            & 16x8x8        & ReLU       & kernel size 2, stride 2 \\
        Classifier           & -             & -          & -        \\
        MLP                  & 16x4x4,120    & ReLU       & -        \\ 
        MLP                  & 120,84        & ReLU       & -        \\ 
        MLP                  & 84,10         & -          & Softmax   \\
        \bottomrule
    \end{tabular}
\caption{Neural module -- LeNet5 for MNIST-Addition experiments. \label{tab:lenet-mnist}}
\end{table}

\begin{table}[ht!]
\centering
    \begin{tabular}{ccccc}
        \toprule
        Variables            & Width    & Height    & Number of Pieces   & Class count  \\
        784                  & 28       & 28        & [4,7,28]           & 10        \\ \bottomrule
    \end{tabular}
    \caption{Probabilistic Circuit module -- EiNet for MNIST-Addition experiments. \label{tab:einet_mnist}}
\end{table}

\begin{table}[ht!]
\centering
    \begin{tabular}{cccc}
        \hline \hline
        Type                 & Size/Channels & Activation & Comment  \\ \hline \hline
        Conv 5 x 5           & 32            & ReLU       & stride 1 \\ \hline
        Conv 5 x 5           & 32            & ReLU       & stride 1 \\ \hline
        Conv 5 x 5           & 32            & ReLU       & stride 1 \\ \hline
        Conv 5 x 5           & 32            & ReLU       & stride 1 \\ \hline
        Position Embedding   & -       & -          & -        \\ \hline
        Flatten              & axis: [0, 1, 2 x 3]             & -          & flatten x, y pos.    \\ \hline
        Layer Norm           & -             & -          & - \\ \hline
        MLP (per location)               & 32           & ReLU          & -        \\ \hline
        MLP (per location)               & 32           & -          & -        \\ \hline
        Slot Attention Module             & 32             & ReLU          & -    \\ \hline
        MLP               & 32            & ReLU          & -        \\ \hline
        MLP               & 16            & Sigmoid          & -        \\ \hline
    \end{tabular}
    \caption{Baseline slot encoder for ShapeWorld4 experiments. \label{tab:slotShapeWorld4}}
\end{table}

\textbf{ShapeWorld4 Experiments} For the baseline slot attention experiments with the ShapeWorld4 data set, we have used the architecture presented in Tab.~\ref{tab:slotShapeWorld4}. 
For further details on this, we refer to the original work of \citeA{SlotAttention}.
The slot encoder had a number of 4 slots and 3 attention iterations over all experiments. 

For the SLASH Attention experiments with ShapeWorld4, we have used the same slot encoder as in Tab.~\ref{tab:slotShapeWorld4}, however, we replaced the final MLPs with 4 individual EiNets with Poon-Domingos structure \cite{SPNs}. 
Their hyperparameters are represented in Tab.~\ref{tab:einets_ShapeWorld4}.

On CLEVR, we also used the ``bigger'' slot encoder architecture for the CLEVR images as in \cite{SlotAttention} which have higher resolution than the Shapeworld4 images. The PC architecture used is the same for CLEVR, but the number of slots is increased to 10.

\begin{table}[ht!]
\centering
    \begin{tabular}{cccccc}
        \hline \hline
        EiNet     & Variables    & Width   & Height   & Number of Pieces   & Class count  \\ \hline \hline
        Color       & 32           & 8       & 4        & [4]                & 9            \\ \hline
        Shape       & 32           & 8       & 4        & [4]                & 4            \\ \hline
        Shade       & 32           & 8       & 4        & [4]                & 3            \\ \hline
        Size        & 32           & 8       & 4        & [4]                & 3            \\ \hline
    \end{tabular}
    \caption{Probabilistic Circuit module -- EiNet for ShapeWorld4 experiments. \label{tab:einets_ShapeWorld4}}
\end{table}

The learning rate for the baseline slot encoder was 0.0004 and 512. 
The learning rate and batch size for SLASH Attention were 0.01 and 512 for ShapeWorld4 and CLEVR for the PCs, and 0.0004 for the slot encoder.

\subsection{Training times for SLASH Attention}
\label{app:nesy-train-times}

\begin{table}[ht!]
    \centering
    \begin{tabular}{lcccccc}
    \toprule
        & \multicolumn{3}{c}{ShapeWorld4} & \multicolumn{3}{c}{CLEVR}\\
        & Baseline &SLASH & SAME & Baseline &SLASH & SAME\\
        \midrule
         Forward pass &-  &1.7 & 1.7 & -& 85.7& 78.1\\
         Potential Solutions &- &50.1 & 8.8 &-& 140.5 &61.7 \\
         Gradients &-&12.2&11.9 &-& 71.3 & 64.2\\
         Backward pass&-& 32.9& 31.4 &-& 273.1 & 243.2\\
        \midrule
         Sum training & 9.4 &96.9& 53.8 & 95.2 &570.6 & 447.3\\
         \bottomrule
    \end{tabular}
    \caption{Average training times per epoch in seconds. The four training stages as well as the total training time per epoch are listed.}
    \label{tab:time-details}
\end{table}

\begin{figure}[ht!]
    \centering
    \includegraphics[scale=0.21]{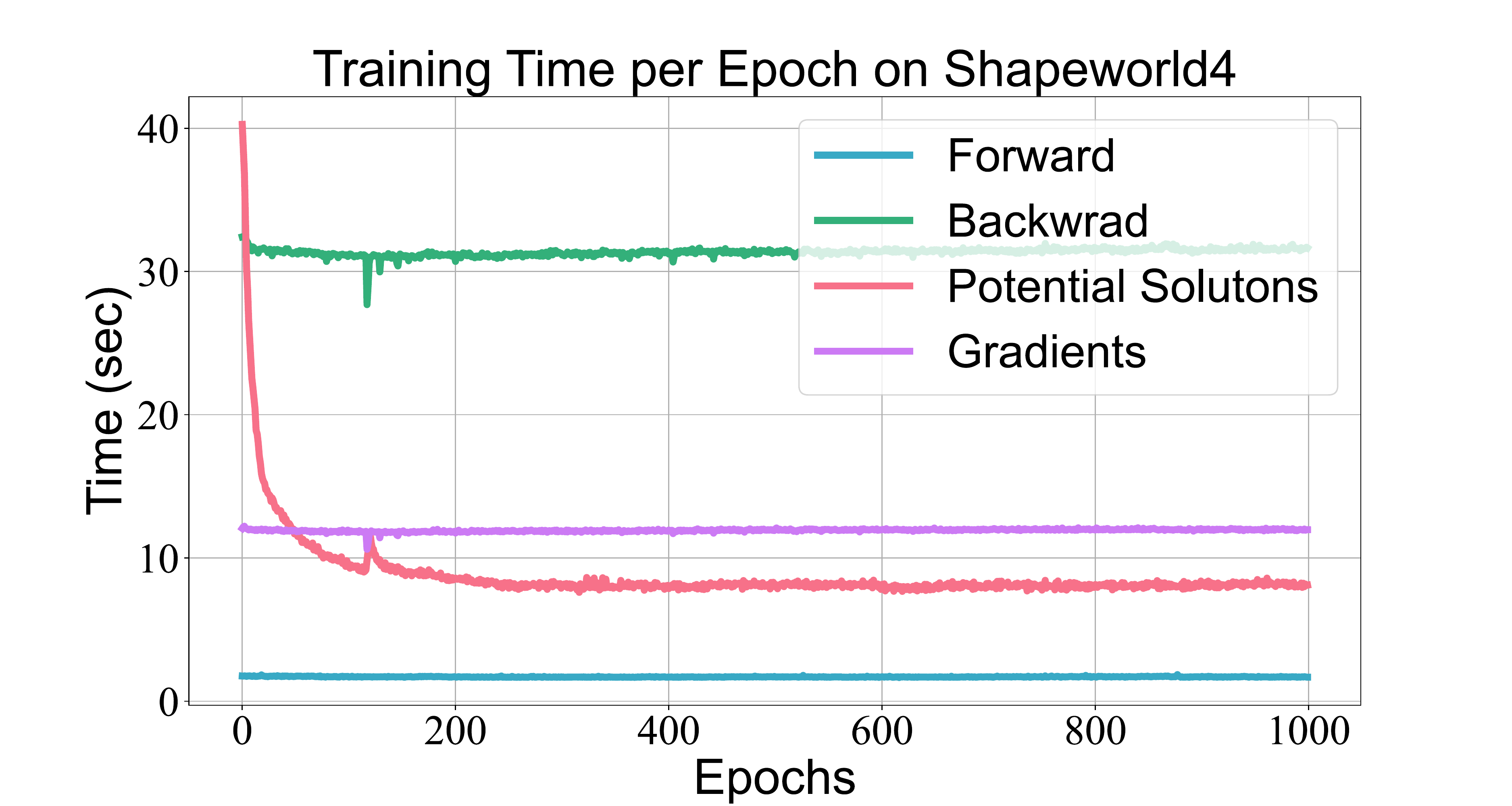}
    \caption{Training time of the four training steps of the SLASH pipeline with SAME. 
    Over time, SAME reduces the time spent on computing Potential Solutions, while all other steps stay constant in time.}
    \label{fig:s4-convergence}
\end{figure}
In Sec.~\ref{sec:object-centric-learning} we saw that there is still some gap between SLASH and its baseline. 
Here we want to have a closer look at where the overhead is coming from.
The training of SLASH can be seen as four steps: The forward pass, computing potential solutions with ASP, computing gradients and lastly the backward pass. 
Tab.~\ref{tab:time-details} gives an overview of the average time spent on each of these steps per epoch. 
The first observation we make is that the forward and backward pass in sum takes longer than the total training of the baseline. 
This is because we are using EinsumNetwork's as the NPPs and that we are using a NPP for each object concept instead of using a single MLP for all concepts and objects at once.
As a result, a lot more parameters are used in total, which increases the time spent on neural computations. 
The biggest bottleneck though is computing the potential solutions, which makes up more than 50\% of the training time.
Fig.~\ref{fig:s4-convergence} shows how SAME helps to mitigate this overhead and reduces the average time to compute Potential Solutions from 49 seconds to 8.8 seconds, making it not longer the training bottleneck.
Computing the gradients stays constant over time and is responsible for 20\% of the total training time for SAME. 
In general, DPPLs as of now utilize a GPU for neural computations, while solving and computing gradients happens on the CPU.
As argued before, this suggests that an interesting research direction would be to find a closer integration of the neural and symbolic components of the pipelines for parallel and faster training.

\clearpage
\bibliography{jair}
\bibliographystyle{theapa}

\end{document}